\pdfoutput=1

\documentclass[11pt]{article}

\usepackage[]{acl2023}

\usepackage{times}
\usepackage{latexsym}

\usepackage[T1]{fontenc}

\usepackage[utf8]{inputenc}

\usepackage{microtype}

\usepackage{amsmath}
\usepackage{amssymb}
\usepackage{booktabs}
\usepackage{graphicx}
\usepackage{subcaption}
\usepackage{xspace}
\usepackage{arydshln}
\usepackage{bbm}
\usepackage[normalem]{ulem}
\usepackage{multirow}
\usepackage{tabularx}
\usepackage{xspace}
\usepackage{physics}
\usepackage[acronym,shortcuts]{glossaries}
\usepackage{fancyvrb}
\VerbatimFootnotes    
\usepackage{soul}

\newcommand{\cs}[1]{\begin{color}{green!50!black}CS: #1\end{color}}
\newcommand{\hy}[1]{\begin{color}{magenta}HY: #1\end{color}}

\title{Implicit Affordance Acquisition via Causal Action–Effect Modeling in the Video Domain}

\author{Hsiu-Yu Yang \\ 
        University of Stuttgart \\
        \small{\texttt{hsiu-yu.yang@ims.uni-stuttgart.de} }\\\And
        Carina Silberer \\
        University of Stuttgart \\ \small{\texttt{carina.silberer@ims.uni-stuttgart.de}} \\}

\begin{document}
\maketitle

\newacronym{cae}{CAE\xspace}{Causal Action--Effect}
\newacronym{vl}{VL\xspace}{visual--linguistic}
\newacronym{prost}{PROST\xspace}{Physical Reasoning about Objects through Space and Time}

\newacronym{multi}{MULTI-CAE\xspace}{Multi-task on Causal Action--Effect Modeling}
\newacronym{mam}{MAM\xspace}{Masked Action Modeling}
\newacronym{mem}{MEM\xspace}{Masked Effect Modeling}
\newacronym{map}{MAP\xspace}{Masked Action Prediction}
\newacronym{mep}{MEP\xspace}{Masked Effect Prediction}

\newcommand{\vl}{visual--linguistic\xspace}
\newcommand{\mamVTten}{\textsc{MAM-VL-10PER}\xspace}
\newcommand{\mamVTtenrandom}{\textsc{MAM-VL$_{Rnd}$-10PER}\xspace}

\newcommand{\mamVT}{\textsc{MAM-VL}\xspace} 
\newcommand{\mamVTrandom}{\textsc{MAM-VL$_{Rnd}$}\xspace} 

\newcommand{\mamT}{\textsc{MAM-L}\xspace} 
\newcommand{\mamTrandom}{\textsc{MAM-L$_{Rnd}$}\xspace} 

\newcommand{\mamVTrandomTnoAFT}{\textsc{MAM-VL$_{Rnd}$-NO-AFT}\xspace} 

\newcommand{\memVTten}{\textsc{MEM-VL-10PER}\xspace}
\newcommand{\memVTtenrandom}{\textsc{MEM-VL$_{Rnd}$-10PER}\xspace}

\newcommand{\memVT}{\textsc{MEM-VL}\xspace} 
\newcommand{\memVTrandom}{\textsc{MEM-VL$_{Rnd}$}\xspace} 

\newcommand{\memV}{\textsc{MEM-V}\xspace} 
\newcommand{\memVrandom}{\textsc{MEM-V}\xspace} 

\newcommand{\multiVT}{\textsc{Multi-CAE-VL}\xspace}
\newcommand{\multiT}{\textsc{Multi-CAE-L}\xspace}
\newcommand{\multiV}{\textsc{Multi-CAE-V}\xspace}
\newcommand{\multiVTrandom}{\textsc{Multi-CAE-VL$_{Rnd}$}\xspace}

\newcommand{\flava}{FLAVA\xspace}
\newcommand{\roberta}{RoBERTa-B\xspace}
\newcommand{\avglm}{AvgLM\xspace}

\newcommand{\words}[1]{``#1''\xspace}
\newcommand{\word}[1]{``#1''\xspace}
\newcommand{\object}[1]{\textsl{#1}\xspace}
\newcommand{\prostobject}[1]{\textit{#1}\xspace}
\newcommand{\prostafford}[1]{\textbf{#1}\xspace}

\begin{abstract}
Affordance knowledge is a fundamental aspect of commonsense knowledge.
Recent findings indicate that world knowledge emerges through large-scale self-supervised pretraining, motivating our exploration of acquiring affordance knowledge from the visual domain.
To this end, we augment an existing instructional video resource  
to create the new \gls{cae} dataset\footnote{\label{caeDataUrl} The project code can be found at \url{https://github.com/Mallory24/cae_modeling}} and design two novel pretraining tasks---\gls{mam} and \gls{mem}---promoting the acquisition of two affordance properties in models: behavior and entity equivalence,  respectively.
 We empirically demonstrate the effectiveness of our proposed
 methods in learning affordance properties. Furthermore, we show that a model pretrained on both tasks outperforms a strong image-based \vl foundation model (\flava) as well as pure linguistic models on a zero-shot physical reasoning probing task. 
\end{abstract}

\section{Introduction}
\label{sec:introduction}Affordances refer to the potential actions and interactions that objects offer/are available to intelligent agents~\cite{Gibson1977-rq}.
For example, a chair affords sitting and a cup affords drinking.
In the context of artificial intelligence, affordance understanding involves training an agent to recognize and interpret affordances in the 
real world, allowing for seamless action anticipation and planning~\citep{Ardon2021-os, DBLP:conf/nips/NagarajanG20}.
Recent advancements in large language models (LLMs) have enabled researchers to use 
their extensive everyday knowledge to decompose high-level natural language instructions into low-level actions for embodied agents  
\citep{DBLP:conf/corl/SayCan_Ichter22, DBLP:conf/corl/Inner_mono_Huang22}.
The lack of physical grounding, however, limits these models to understand affordances \cite{bisk-etal-2020-experience}. 
\begin{figure}
    \centering
    \resizebox{0.5\textwidth}{!}{\includegraphics{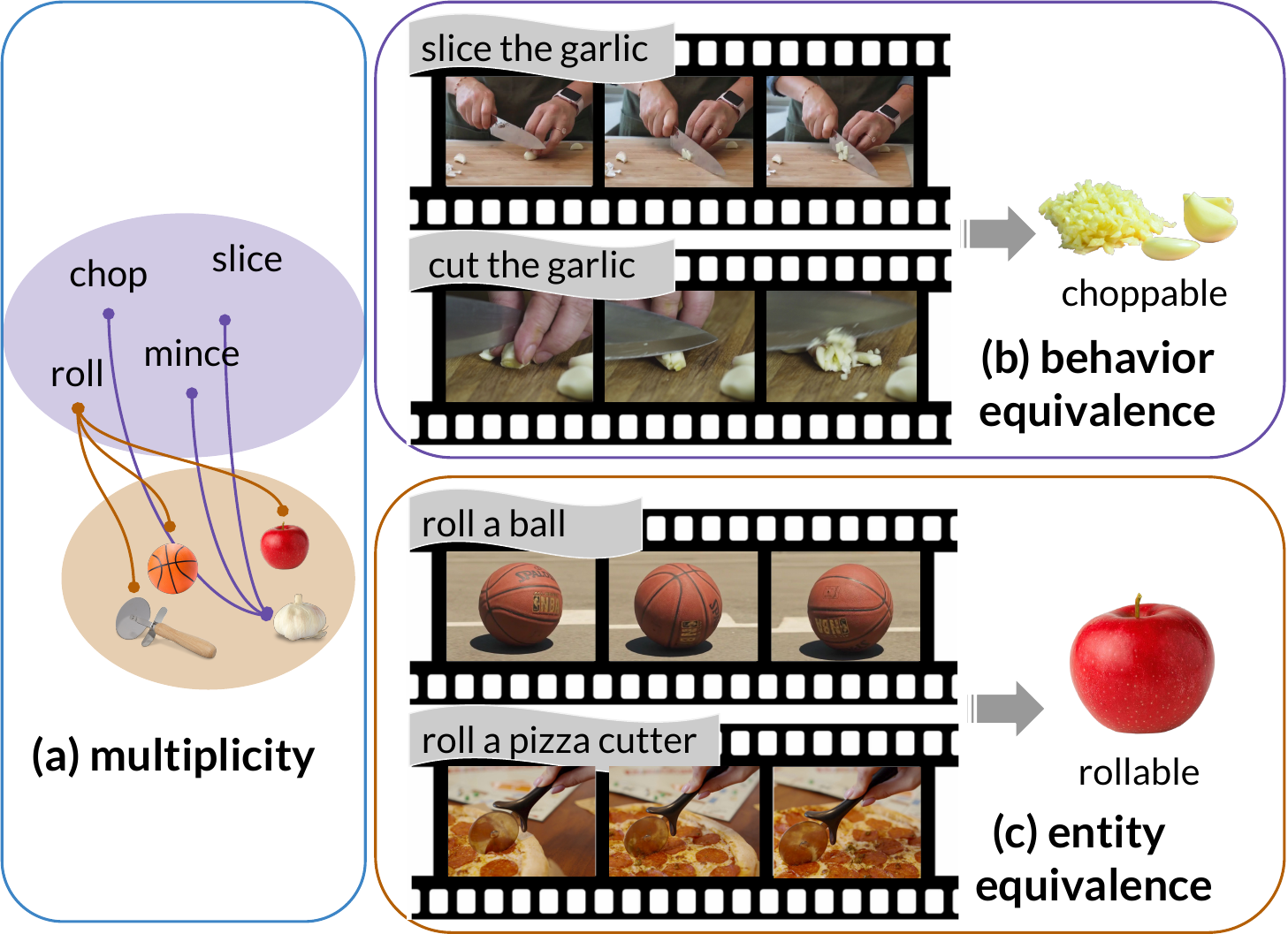}
    }
    \caption{We address the (a) multiplicity issue of affordance learning by modeling causal action--effect during 
    pretraining 
    to implicitly induce two essential properties of affordance: (b) behavior 
    and (c) entity equivalence.}
    \label{fig:intro}
\end{figure}
The main challenge in learning affordance is multiplicity. 
Put simply, different objects of 
distinct semantic classes 
potentially facilitate the same action, and a single object can offer multiple feasible actions~\cite{LuPAD-L}.
As illustrated in Figure~\ref{fig:intro}(b) \& (c), both the \word{ball} and \word{pizza cutter} can afford the action \word{roll} while garlic supports actions like \word{slide} and \word{cut}.
More precisely, two principles serve as the foundation for realizing multiplicity~\cite{Sahin2007-pq}.
The first is \textit{behavior equivalence}, which states that different actions can produce the same effect on a given object, e.g., \word{slice}, \word{cut} and \word{chop} can all make \word{garlic} into smaller pieces (Fig.~\ref{fig:intro}(b)). 
The second, \textit{entity equivalence},  suggests that executing the same action on different objects can lead to identical outcomes, e.g., rolling an \word{apple} producing a similar motion change to rolling the \word{ball} (Fig.~\ref{fig:intro}(c)).
Learning such associations poses a significant challenge for generalizing knowledge to novel objects and unfamiliar scenarios~\cite{LuPAD-L}.

Existing methods address the generalization problem either by identifying shared characteristics among objects of an affordance category, or iteratively reinforcing cross-modal representation consistency \cite{LuPAD-L, DBLP:conf/ijcai/LuoZ00T21}.
However, these approaches are devised for specific affordance tasks in supervised settings, raising questions about their adaptability to other tasks \cite{DBLP:conf/aaai/PIQA_Bisk, aroca-ouellette-etal-2021-prost}.
Given recent findings on the emergence of world knowledge from large-scale self-supervised pretraining \cite{petroni-etal-2019-language} and the importance of the visual domain for distilling certain knowledge~\cite{shwartz-choi-2020-neural, paik-etal-2021-world}, we study the implicit acquisition of affordance knowledge through \vl (VL) pretraining.
As we seek for a high diversity in action and effect categories, we explore affordance learning in a noisy real-world video domain~\cite{DBLP:journals/corr/Ebert_verb_trajectory}.
To this end, we leverage step-by-step instructional videos accompanied by subtitles~\cite{DBLP:conf/iccv/MiechZATLS19}, from which we extract affordance-relevant clip--subtitle pairs.
The resulting dataset, referred to as \gls{cae}{(\textbf{C}ausal \textbf{A}ction--\textbf{E}ffect Dataset)}, contains $4.1$M clip--subtitle pairs for modeling diverse actions and their effects.
We then introduce two pretraining tasks---\gls{mam} and \gls{mem}---to induce behavior equivalence and entity equivalence, respectively.

To validate our approach, we conduct intrinsic evaluations addressing two research questions: \textbf{(RQ1)} can models trained with \gls{mam} and \gls{mem} adequately learn fundamental principles of affordances?
\textbf{(RQ2)} what are the benefits of joint task training?
We then assess the encoded affordance knowledge in the grounded representations to answer the third question: \textbf{(RQ3)} how effective is our causal action--effect pretrained model on solving an affordance probing task?

\section{Related Work}
\label{sec:related_work}\paragraph{Affordance Learning.}
Several works mine affordances from text corpora by identifying semantically plausible verb--object pairs \cite{loureiro-jorge-2018-affordance, Persiani2019-hd, Chao2015-kh}. Closely related are selectional preferences \cite{Pantel2007-ci, Erk2007-wl}, i.e., typical arguments (e.g., objects) of a verbal predicate. 
Another line of research targets \textit{visual affordances}, their detection and categorization in visual input \cite{DBLP:journals/csur/HassaninKT21}. 
Affordances underlie the multiplicity property, i.e.,~multiple objects can be mapped to one affordance category and vice versa, making supervised learning challenging. Few works address this by enhancing the multimodal representation consistency iteratively \cite{LuPAD-L} or by finding joint object features of a given affordance class \cite{DBLP:conf/ijcai/LuoZ00T21}.
Inspired by the robotics domain \cite{Sahin2007-pq,Dag2010-hd, Dehban2016-ok, Jaramillo-Cabrera2019-oh}, we model the relationships between actions, objects, and the observed effects for encoding affordance knowledge implicitly via a self-supervised setup.  
In contrast to 
\citet{merullo-etal-2022-pretraining} that model object trajectories as effects in a closed simulated environment, we use action--object--effect relations mined from diverse web-crawled instructional videos. 
\paragraph{Causal Modeling of Action Verbs. }
Grounding the meaning of action verbs to the state changes of the manipulated objects is gaining attention in the NLP community 
\cite{RACsurvey}.
\citet{Gao2016-tj} are the first 
to explore causal verb modeling  
in the cooking domain \cite{regneri-etal-2013-grounding} for 
grounded semantic role labeling. 
\citet{Bosselut2017-aj} use symbolic action--effect modeling for procedural text understanding. 
\citet{Gao2018-kc} 
introduce 
multimodal action--effects prediction---linking 
action verbs 
to their 
effects in static images. 
Several works approach action--effect modeling in a simulated environment. 
\citet{zellers-etal-2021-piglet} collect action and object state transitions in AI2-THOR \cite{AI2-THOR_Kolve}, and ground language models to the physical world through symbolic representations (e.g.,~\texttt{isWarm=True}). 
\citet{hanna-etal-2022-act} address effect prediction in AI2-THOR in a more challenging setup where the image showing the post-action is to be chosen, while  
\citet{dagan-etal-2023-learning} predict change labels from the visual input. 
Our goal, in contrast, is to implicitly augment models with affordance knowledge through causal action--effect modeling. 
Moreover, we do not bound object states to a fixed set of categorical labels, but exploit the temporal dimension to represent perceptual effect changes.

\section{The CAE Dataset}
\label{sec:datasets}We base our work on the hypothesis that affordance knowledge can emerge from large-scale self-supervised pretraining. 
We choose to build upon HowTo100M \cite{DBLP:conf/iccv/MiechZATLS19}, 
the largest and most diverse instructional video dataset available at the time of writing
\cite{DBLP:conf/cvpr/COIN_TangDRZZZL019, DBLP:conf/cvpr/CrossTask_ZhukovACFLS19, YT_Kuehne2019}.
It contains $136$M~weakly-paired
As our interest lies in modeling causal action--effect, we employ a series of automatic procedures to extract useful video clips from HowTo100M: 
(1) identify a set of result verbs by leveraging various linguistic resources
(2) locate casual action--effect video clips via parsing subtitles 
to match the result verb set. 
We call our resulting clip--subtitle pairs the \gls{cae} dataset.\footref{caeDataUrl}
\subsection{Identify Result Verbs}\label{subsec:result_verbs}
To model perceptual causal change of actions, we are interested in a specific verb type called result verbs. 
Result verbs
cause state changes on their arguments, including changes in volume, area, gradable scale, and motion \cite{Levin_undated-pc}. 
One can reuse an existing collection of result verbs from \citet{Gao2018-kc}; yet, it has limited coverage ($62$ verb classes, of which~$39$ are covered in our list). 
Furthermore, to facilitate a reproducible and systematic method for identifying result verbs, we define $2$ 
criteria that a potential result verb should meet: 
(1) \textbf{visualness}: it can be visually perceivable (2) \textbf{effect-causing}:
it can cause physical results on the object it acts upon. 
We use several semantic  resources, including VerbNet \cite{kipper-etal-2006-extending} for its informative selectional restrictions, and imSitu  \cite{DBLP:conf/cvpr/YatskarZF16}, to leverage its exhaustive frame--semantic annotations on visual verbs.
Moreover, we employ FrameNet (FN; \citealp{baker-etal-1998-berkeley-framenet}) to extract
(undisambiguated) situational frames a verb could evoke 
and use the associated frame elements
to identify 
potential result verbs.\footnote{See 
    App.~\ref{app:result_verbs} for details on the versions we use. }
We will further use the extracted frames for generalization analysis on unseen verb classes (detailed in Sec.~\ref{sec:experimental_setup}).
To automate result verb identification,
we define
heuristics to verify \textbf{visualness} and \textbf{effect-causing} properties.
\begin{figure}
    \centering
    \resizebox{0.5\textwidth}{!}{
    \includegraphics{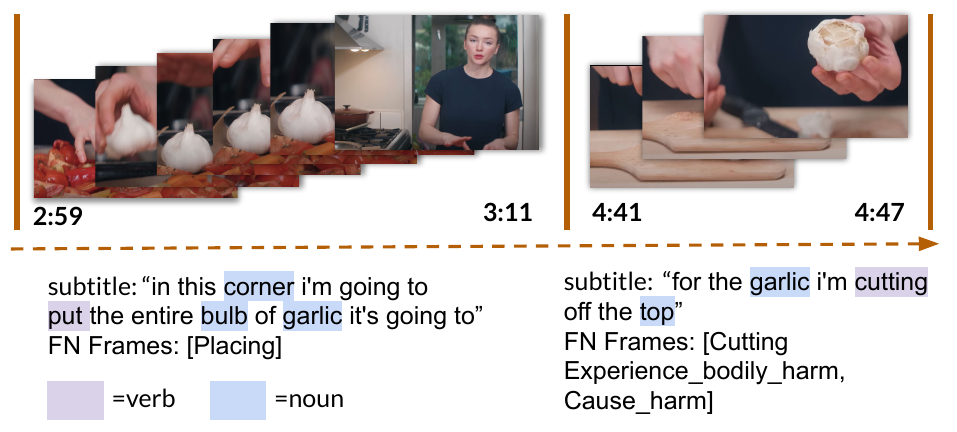}
    }
    \caption{\gls{cae} Clip--subtitle pairs.
    }
    \label{fig:cae}
\end{figure}
\paragraph{Visualness.}
We retain potential result verbs from imSitu's visually perceivable verbs, whose semantic role in the second position is valid, e.g.~\texttt{Item}.\footnote{A list of invalid roles are detailed in App.~\ref{app:result_verbs}} 
From VerbNet, we extract visual verbs
based on the selectional restrictions on the thematic role of its verb class. 
Precisely, if a verb class, e.g., \textsl{spank-18.3}, specifies the \texttt{Patient} role to be concrete or solid, we consider its grouped verb members to possess visualness, e.g., \words{whisk} and \words{whip}. 

\paragraph{Effect-causing.}
To automatically 
determine
the effect-causing characteristic of a verb, we check if the thematic role combination of its 
verb class on VerbNet follows (\texttt{Agent}, \texttt{Patient}, \texttt{Result}), e.g., the verb \words{split} in the verb class \textsl{break-45.1} is confirmed.
We then cross-check with FN on the potential result verbs that passed the previous visualness test and the initial effect-causing test to ensure this property. 
Specifically, we check if a candidate result verb can evoke any frame that contains either the \texttt{Result} or the \texttt{Effect} frame element, e.g.,  the verb \words{simmer} evokes the \textsc{Apply\_heat} frame, and has the \texttt{Result}  element.
To consolidate the \textbf{visualness} and the \textbf{effect-causing} information obtained from the various lexical resources, we merge the verbs on their lemma, 
 e.g.,\words{grill} is judged as a result verb because 
its verb sense \textsl{grill-45.3}  captures visualness and effect-causing properties, even though 
the other, \textsl{grill-26.3-2}, lacks such information on VerbNet. 
We derive in total $\textbf{236}$ sure cases with 
both \textbf{visualness} and \textbf{effect-causing} characteristics (an overview of sure and unsure cases can be found in~Tab.~\ref{tab:sure_unsure_result_verbs} of App.~\ref{app:result_verbs}), 
which we use to locate causal action--effect video clips 
described in the following.
\begin{table}[t]
\small
\resizebox{0.45\textwidth}{!}{
\begin{tabular}{@{}llll@{}}
    \toprule
    {} &     Videos &  Clips & Top 5 Result Verbs \\
    \midrule
    food      &  167k &  2.15M & make, put, cook, mix, cut \\
    hobbies   &   60k &   0.8M & make, put, cut, pull, fold \\
    cars      &   12k &   0.1M & make, put, pull, turn, push \\
    pets      &    8k &   90k & make, put, cut, set, build \\
    sports    &    3k &   39k & make, put, cut, pull, build \\
    \bottomrule
    \end{tabular}
    }
    \caption{Number of unique videos, video clips, and the top 5 result verbs across $5$ selected video domains.
    }\label{tab:cae_5_domains}
\end{table}
\subsection{Extract CAE Video Clips}\label{subsec:cae_dataset}
Given the large size of the HowTo100M dataset, we 
down-sample the video pool with several heuristics 
and focus on $13$~video domains with a high density of unique result verbs (see App.~\ref{app:subtitles} for details).
To extract video clips from this pool relevant for causal action--effect modeling, we rely on the paired subtitles and their 
linguistic information such as PoS tags and dependency labels. 
Particularly, we only keep clips 
with a single occurrence of 
one of our results verbs.  
To mitigate information leakage 
due to overlapping video frames, we 
enforce a minimum 
difference of $5$~seconds between adjacent clips 
at the per-video level.\footnote{We discard consecutive clips as they usually capture redundant information, i.e.,~the same result verb.}  
In addition, to annotate proxy objects (i.e., nouns), we find the set of objects (within a subtitle) by considering the intersection of nouns labeled with \texttt{dobj} or \texttt{pobj} relations  
and with high concreteness ratings \mbox{(i.e., > $4$}; \citealp{brysbaert_2014}). 
Through this process, we derive in total 
$\textbf{4.1}$M~video clips with $\textbf{235}$~unique result verb types.\footnote{Only the phrasal verb \textit{warm\_up} is not considered.} 
Figure~\ref{fig:cae} gives examples (the comparison to the underlying  HowTo100M clips can be found in App.~\ref{app:subtitles}, Fig.~\ref{fig:cae_howto100m}).\footnote{To align with our visual feature extraction procedure, the shown timestamps are extended $3$~seconds before the original timestamp (see App.~\ref{app:model}). Note the released \gls{cae} dataset contains the original timestamp.} 
As Table~\ref{tab:cae_5_domains} shows, most of the verbs are
prevalent across domains, e.g.~\word{make}. 
Since our goal is to maintain a naturalistic dataset, we keep such 
common 
verbs (see App.~\ref{app:subtitles} for details.).
Recall that we built upon noisy weakly-paired clip--text data, resulting in many items without visual occurrences of both our target actions and objects ($70\%$~of $288$~\gls{cae} samples analyzed by a postgraduate student). 
We keep these ``background examples'' as a preliminary study  \cite{DBLP:journals/corr/KuehneYT19}.

\section{Causal Action--Effect Modeling}
\label{sec:method}
\begin{figure*}
    \centering
    \scalebox{0.6}{%
    \includegraphics{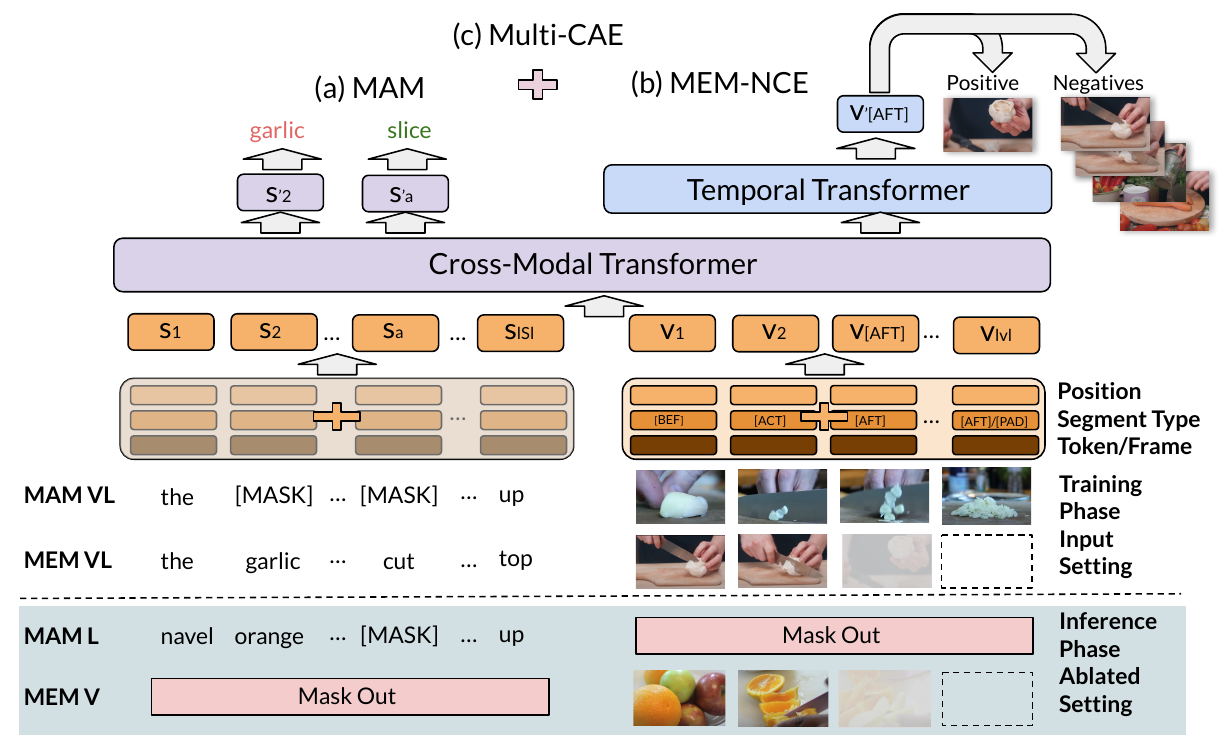}
    }
    \caption{Our proposed pretraining tasks on the hierarchical video--language architecture HERO \cite{li-etal-2020-hero}:
    (a) MAM for behavior equivalence on the output of Cross-Modal Transformer (Sec.~\ref{subsec:map}) (b) MEM for entity equivalence on the output of Temporal Transformer (Sec.~\ref{subsec:mep}). (c) MULTI-CAE for joint-task training. During inference (blue-gray background), we also ablate one modality via attention masks for a systematic intrinsic evaluation.
    }
    \label{fig:model}
\end{figure*}
Our goal is to induce the basic principles of affordance knowledge, i.e., behavior equivalence and entity equivalence
(Fig.~\ref{fig:intro} of Sec.~\ref{sec:introduction}).
We design two pretraining tasks: \gls{mam} and \gls{mem}, and further explore the benefit of joint task training by alternating task-specific samples to optimize the model, referred to as \gls{multi}. In order to gauge the understanding of affordance properties, the corresponding intrinsic tasks called \gls{map} and \gls{mep} are introduced.
We build upon the existing video--language hierarchical framework HERO \cite{li-etal-2020-hero}, consisting of a \textbf{Cross-Modal Transformer} to learn contextualized embeddings between a subtitle and a video clip locally and a \textbf{Temporal Transformer} to learn video embeddings on the global context (i.e, the whole video clip). 
Figure~\ref{fig:model} illustrates the overall architecture and the pretraining tasks  (for model details see  App.~\ref{app:model}).
\subsection{Masked Action Prediction (\gls{map})}\label{subsec:map}
\paragraph{Task Definition.} 
Our underlying assumption is that a video clip of the \gls{cae} dataset visually depicts the change of the pre-condition (\texttt{[BEF]}), the action process (\texttt{[ACT]}), and the post-condition (\texttt{[AFT]}) in sequential order throughout an action execution \cite{RACsurvey}. 
Given a \gls{cae} clip--subtitle pair, 
 where the action~\mbox{$a \in A$} (a result verb) 
 is masked. The task is to predict verb~$a$. 

\paragraph{Masked Action Modeling (\gls{mam}).}
We address \gls{map} on the local context, i.e.,~the contextualized multimodal embeddings computed by the Cross-Modal Transformer (Fig.~\ref{fig:model} (a)~MAM).
To prepare the multimodal inputs, we extract a series of video frames~\mbox{$V = \{\vb{v}_j\}_{j=1}^{|V|}$} from the video clip every 2 seconds, and tokenize the corresponding subtitle into a sequence of tokens \mbox{$S = \{\vb{s}_i\}_{i=1}^{|S|}$} (refer to App.~\ref{app:model} for details). We replace the (masked) target verb with the special \texttt{[MASK]} token.

\paragraph{Training.}   
During pretraining, we seek to encourage the model to learn holistic semantics. 
We thus mask the verb~\mbox{$a$} and each word in \mbox{$S$} with a chance of~$15\%$.\footnote{Following BERT, 80\% and 15\% of the target verbs are replaced with \texttt{[MASK]} and a random token, respectively, 5\% of them are unchanged.}

We empirically verify the benefit of this masking strategy during preliminary experiments (see App.~\ref{app:mam} for details). 
The objective is to reconstruct \mbox{$\vb{s}_{a^*}$} (the action verb and some random words) based on the observation of unmasked tokens \mbox{$S_{\setminus \vb{s}_{a^*}}$} and the video clip $V$ such that the learnable parameters \mbox{$\theta$} are updated with the loss function:
    \begin{gather*}\label{eqn:mam}
    \mathcal{L}_{MAM}(\theta) = -\log P_{\theta}(\vb{s}_{a^*} | S_{\setminus \vb{s}_{a^*}} \| V)
    \end{gather*}

\paragraph{Inference.} 
During inference, only the target verb~\mbox{$a$} is masked in the input. 
We feed the preprocessed input to HERO, and consider the token with the highest logits in the \gls{mam} head's output as the predicted token~\mbox{$\hat{a}$}. 

\subsection{Masked Effect Prediction  (\gls{mep})}\label{subsec:mep}
\paragraph{Task Definition.}
Recall from \gls{map} (Sec.~\ref{subsec:map}), the effect of an action on the object is assumed to be visually perceivable in a \gls{cae} clip--subtitle pair. 
The goal of \gls{mep} is to predict the masked  \texttt{[AFT]} (post-condition) subclips in a discriminative
setup:  
for each  \texttt{[AFT]} subclip (frame), the correct frame from candidate frames with the same \textbf{video id} as the given video clip is to be selected.\footnote{The rest the of intra-video clip \texttt{[AFT]} frames are excluded from the candidate set (see App.~\ref{app:mem} for details).}

\paragraph{Masked Effect Modeling (\gls{mem}).}
Unlike \gls{map}, the task is addressed globally 
with the temporal contextualized embeddings computed by the Temporal Transformer (Fig.\ref{fig:model} (c)~MEM). 
    Given a preprocessed pair of video frames~\mbox{$V = \{\vb{v}_j\}_{j=1}^{|V|}$} and subtitle tokens~\mbox{$S = \{\vb{s}_i\}_{i=1}^{|S|}$}, we arbitrarily divide the video clip into three parts in a near-equal manner: \texttt{[BEF]} (pre-condition), \texttt{[ACT]} (action), \texttt{[AFT]} (post-condition)\footnote{We leave automatic action localization \cite{DBLP:conf/eccv/ActionFormerZhangWL22} to determine effect frames more reliably, to future work.}
    and mask the video frames~\mbox{$\vb{v}_{\texttt{[AFT]}}$}
    corresponding to the \texttt{[AFT]} post-condition subclips with zero-vectors.

\paragraph{Training. }
     During pretraining, the objective is to reconstruct~\mbox{$\vb{v}_{\texttt{[AFT]}}$} given the observation of unmasked video frames~\mbox{$V_{\setminus \vb{v}_{\texttt{[AFT]}}}$} and  subtitle~\mbox{$S$}:
    \begin{gather*}\label{eqn:mem}
        \mathcal{L}_{MEM}(\theta) = -\log P_{\theta}(\vb{v}_{\texttt{[AFT]}}| V_{\setminus \vb{v}_{\texttt{[AFT]}}} \| S)
    \end{gather*}
    Concretely, following \citet{li-etal-2020-hero}, we optimize the model with a contrastive loss by employing the softmax version of the NCE loss function \cite{NCEsoftmax},
    thus, 
        \begin{gather*}
        \small
        \mbox{$P_{\theta}(\vb{v}_{\texttt{[AFT]}}| V_{\setminus \vb{v}_{\texttt{[AFT]}}} \| S)$} \approx \frac{\exp(\vb{v}^\prime_{\texttt{[AFT]}} \cdot \vb{v}_{\texttt{[AFT]}} / \tau)}{\sum_{i=0}^C \exp(\vb{v}^\prime_{\texttt{[AFT]}} \cdot c_i / \tau)}
        \end{gather*}
    where~\mbox{$\tau$} is a temperature to control the strength of the penalties on the negative samples, 
    and where the global contextualized embeddings~\mbox{$\vb{v}^\prime_{\texttt{[AFT]}}$}, computed by the Temporal Transformer, are taken to compute the similarity to each video frame ~\mbox{$c_i$} in ~\mbox{$C$}, a candidate set comprising the correct~\mbox{$\vb{v}_{\texttt{[AFT]}}$} and negative video frames sampled from the same \textbf{video id} as the input video clip~\mbox{$V$}.\footnote{In a preliminary study on various negative sampling strategies, we found this strategy most beneficial for  teaching fine-grained effect changes  (see App.~\ref{app:mem}).}

\paragraph{Inference. }
Identical to training, we mask  post-condition video frames  $\vb{v}_{\texttt{[AFT]}}$ 
for reconstruction. 
By using the trained MEM head to compute the dot product between $\vb{v}^\prime_{\texttt{[AFT]}}$ and each of the candidate  frames from the same video id, the highest score obtained is considered as the model's prediction.

\section{Experiments}
\label{sec:experimental_setup}
We first evaluate our proposed methodology for multimodal affordance learning intrinsically on \gls{map} and \gls{mep}, and then on the task-specific probing benchmark \gls{prost}.
\subsection{Intrinsic Evaluation Setup}
\label{subsect:intrinsic_eval_setup}
The intrinsic evaluation (see Sec.~\ref{subsec:map} \& \ref{subsec:mep} for the task descriptions) 
addresses \textbf{(RQ1)} \& \textbf{(RQ2)} (see Sec.\ref{sec:introduction}). Precisely, we assess these two affordance principles by \textbf{(1)} quantifying their generalization ability in a zero-shot framework, i.e., evaluate how well a model performs on unseen action (verb) and object (noun) classes; \textbf{(2)} examining the generalization ability from the perspective of event  \cite{baker-etal-1998-berkeley-framenet} and  lexico-taxonomic \cite{miller-1994-wordnet} semantics.
We thus implement an algorithm controlling the splitting of the \gls{cae} dataset.\footnote{The algorithm is reproducible, allowing researchers to customize their own splits based on their requirements.}

\paragraph{Dataset Split.}
\begin{table}[t]
    \small
    \centering
    \begin{tabular}{@{}lcccc@{}}
        \toprule
        & \multicolumn{1}{c}{video}  
        & \multicolumn{3}{c}{\% zero-shot} \\
        & \multicolumn{1}{c}{clips}
        & videos
        & actions 
        & objects  
        \\
        \midrule
     train & 2,818,433 & -- & -- & --    \\ 
     val  & 646,423 & 1.07 & 45.45 & 0.90  \\
     test & 644,986 & 1.10 & 45.47 & 0.86  \\ \midrule
     all & 4,109,842 & -- & -- & --  \\
     \bottomrule
    \end{tabular}
    \caption{Number of video clips and $\%$ of zero-shot samples w.r.t. the train set in terms of unique video ids, action (verb) classes, object (noun) types. 
    }\label{tab:cae_split}
\end{table}

To examine the action generalization ability on the level of event semantics in a \textbf{generalized zero-shot} setting \cite{Xian2017-oi}, we assign verbs either to seen or unseen classes within a FrameNet frame and remove the restriction that all the verb classes of the test set should be unseen. 
In total, there are $\textbf{143}$ seen and $\textbf{92}$ unseen verb classes, resp. 
As shown in Table~\ref{tab:cae_split}, the train/dev/test split ratio follows ~\mbox{$70\%/15\%/15\%$} and nearly $50\%$~of instances in the dev/test set contain unseen actions (verbs) (see App.~\ref{app:split_stats} for relevant frequency statistics for verbs and verb--noun combinations in train/test). 
Note that the test set is not manually annotated by humans and serves solely as a reference set \cite{YT_Kuehne2019}. 

\paragraph{Model Ablations.}
To ensure a faithful assessment of the generalization ability, we diverge from \citet{li-etal-2020-hero} and use randomized weights rather than \roberta's pretrained weights to initialize HERO's Cross-Modal Transformer.\footnote{According to \citet{li-etal-2020-hero}, RoBERTa's pretrained weights ($12$ layers) are partially taken to initialize the Cross-Modal Transformer ($6$ layers) of HERO.} 
The models pretrained with \gls{mam}, \gls{mem}, and \gls{multi} tasks are referred to as \textbf{\mamVTrandom}, \textbf{\memVTrandom}, and \textbf{\multiVTrandom}, respectively. 
Moreover, to see if the modalities provide complementary information,\footnote{The action is verbally mentioned (see Sec.~\ref{subsec:cae_dataset}.)} 
 we train model variants with one ablated modality
for comparison: \textbf{\mamTrandom} and \textbf{\memVrandom}.  
We do this through
zero-masking on the inter-attention mask. In other words, 
\textbf{\mamTrandom} and \textbf{\memVrandom} are tasked to reconstruct the masked textual/visual target tokens under their own respective unimodal input (see Appendix~\ref{app:pretraining_details} Ablated Models).
During inference, we perform a sanity check, where we ablate one modality to see whether the linguistic/visual context is sufficient to solve the task as illustrated in Figure~\ref{fig:model}. 
See Appendix~\ref{app:pretraining_details} for 
the  pretraining details and the hyperparameters we adopted. 

\paragraph{Metrics.} For \gls{map}, we report the macro-average accuracy on seen/unseen verb classes and their harmonic mean
\cite{Xian2017-oi}, along with the micro-average accuracy.
For the \gls{mep} task, we report the micro-average accuracy, which measures the correctness of predicting all masked \texttt{[AFT]} frames per instance.

\subsection{Probing Task: 
PROST}\label{subsect:extrinsic_eval_setup}

To address \textbf{(RQ3)} (Sec.~\ref{sec:introduction}), we assess the encoded affordance knowledge in our causal action--effect pretrained models by performing a zero-shot evaluation on the \gls{prost} task \cite{aroca-ouellette-etal-2021-prost}.
It intends to probe models on physical commonsense knowledge in a textual cloze-style format, including 
6~affordance concepts: \prostafford{stackable, rollable, graspable, breakable, slidable} and \prostafford{bounceable} for $38$~object classes in total. 
Each of them comprises two templates: (1) the \textbf{original template} asks the model to choose the object among $4$~objects that affords a given action, while 
(2) its \textbf{inverse} 
asks for the object that cannot afford the action; 
e.g.,~\prostobject{eggs} is the non-stackable object in [\prostobject{eggs, books, blocks, boxes}].
The distribution of \textbf{correct answer} positions is uniform across the test instances. Therefore, a model that is found robust exhibits similar effectiveness on the original and its inverse template, as well as on all answer positions.
\begin{table*}[t]
    \centering
    \small
    \begin{tabular}{@{~}l@{~}|l|rrrr||r|r@{~}}
    \toprule 
    Test Set Input & Model  & Seen & Unseen & HM & Micro  & FN frame & Co-Hypo \\
    \midrule
    \multirow{4}{*}{Multimodal} 
    & Majority baseline  & $0.7$ & $1.1$ &  \textbf{$2.0$} & $14.6$  & - & -\\
    & \multiVTrandom  & 23.6 & 13.5 & 17.2 & 30.9 & 16.6 & \underline{36.1} \\ 
    & \mamVTrandom  & $\textbf{\underline{26.8}}$ & $\textbf{\underline{14.9}}$ & $\textbf{\underline{19.1}}$ & $\textbf{\underline{31.7}}$ & \textbf{17.5} & 34.5 \\ 
    \midrule
    \multirow{4}{*}{Language} 
    & \roberta$^\dagger$  & 7.6 & 5.2 & 6.2 & 13.3 & 3.7 & 8.8 \\ 
    & \multiVTrandom  & 15.1 & 10.2 & 12.2 & 19.6 & 11.7 & \textbf{39.0} \\
    & \mamVTrandom  & $16.5$ & $10.6$ & $12.9$ & $19.7$ & \underline{12.6} & 36.6 \\ 
    & \mamTrandom  & ${\underline{18.4}}$ & ${\underline{14.4}}$ & ${\underline{16.2}}$ & ${\underline{23.3}}$ & 10.2 & 37.3\\ 
    \bottomrule
    \end{tabular}
    \caption{
    \textbf{Left}: Results (\%) for MAP on (1) Multimodal (video + subtitles) and (2) Language (subtitles) test input for Seen/Unseen verbs.
    HM = Harmonic Mean, Micro = micro-average accuracy. 
    \roberta$^\dagger$ is a linguistic baseline. 
    \textbf{Right}: Proportion (in \%)
    of false predictions on the set of unseen verb classes, which generalizes in terms of event and lexico-taxonomic semantics, measured through a shared FrameNet~(FN) frame and direct Co-Hyponymy, respectively.
    {Best results} in \textbf{boldface}, best result in each input mode \underline{underlined}. \label{tab:final_map_and_useen_verb_results}}
\end{table*}

\paragraph{Models.}

We 
compare pure language models (LM) against variants of \mamT, \mamVT, and \multiVT, whose textual encoders are initialized with the pretrained \roberta model  (Hugging Face, \citealp{wolf-etal-2020-transformers}) before \gls{cae} pretraining.  
Additionally, to test our hypothesis that the video domain better captures action--effects due to the temporal dimension, we compare against the image-based \vl foundation model~\flava \cite{DBLP:conf/cvpr/FLAVA22}, known for its strong unimodal and multimodal representations. 
Note that before undergoing multimodal pretraining, the textual encoder of \flava is initialized with unimodal pretrained weights,\footnote{The textual encoder is pretrained using the MLM objective on CCNews and BookCorpus.} making its linguistic understanding comparable to pretrained text-based models. 
The models thus all have some prior linguistic knowledge, allowing us to focus on the effect of \gls{cae} pretraining.
Regarding the LMs, we report results for \roberta, and the average (coined \avglm) of the best results  that \citet{aroca-ouellette-etal-2021-prost} report for each of their LMs (GPT, GPT2, BERT, RoBERTa, ALBERT~V2).\footnote{Following  
\citet{aroca-ouellette-etal-2021-prost}, we exclude the not directly comparable models T5 (it does not cover \prostafford{slide}), and UnifiedQA (fine-tuned on task-specific text data).}

For inference, all models are fed \textit{textual input} only, and we follow \citeauthor{aroca-ouellette-etal-2021-prost}'s  (\citeyear{aroca-ouellette-etal-2021-prost})  procedure to obtain the probabilities of each candidate, based on the logits of pretrained MLM (\roberta \& \flava) and MAM heads (ours).

\section{Results}
\label{sec:intrinsic_results}
\subsection{Intrinsic Tasks: \gls{map}}\label{subsect:map}
    The accuracy of all models 
    on predicting the correct verb is overall lower on the \textbf{language-input} test set compared against the \textbf{multimodal} set, where they are fed both, subtitles and video clips (Block~$2$ vs.~$1$, Tab.~\ref{tab:final_map_and_useen_verb_results}, left). 
    The best 
    model, \mamVTrandom, drops by \mbox{$-10.3$pp}~accuracy on seen and \mbox{$-4.3$pp}~on unseen verb classes, suggesting  
    %
    that the task and test set is not trivially solvable without visuals. 
    It also indicates that the VL  model has successfully learned to ground verbs by leveraging the visual effect change 
    (cf.~\citealp{sampat-etal-2022-learning}). 
            Indeed, when omitting the \texttt{[AFT]} frames (the visual effect) during inference, the accuracy of \mamVTrandom drops on Seen and Unseen ($-4.5$pp and $-2.3$pp, resp., no table shown).
            But the benefit of the visual modality for the model seems still limited regarding its generalization ability---the difference on Unseen between \mamVTrandom and its ablated variant \mamTrandom (pretrained on subtitles), is minimal ($+.5$pp for VL);  
             \multiVTrandom even underperforms 
            \mamTrandom on Unseen  irrespective of the test input.  
    
    \paragraph{Analysis: Generalization Ability.}   
    We analyzed the effect of our pretraining tasks on the generalization ability of verbs on the level of situations/events, i.e.,~semantic frames, and less specific verb senses: 
    We measured the number of cases, in which the reference verb~$\mathbf{{v}}$ and the predicted verb~$\mathbf{\tilde{v}}$,    $\mathbf{{v} \neq \tilde{v}}$,  share the same FrameNet~(FN) frame or the same direct WordNet~(WN) hypernym (co-hyponymy).\footnote{We used the NLTK toolkit \cite{bird-loper-2004-nltk}.}
    Table~\ref{tab:final_map_and_useen_verb_results} (right) shows that \mamVTrandom and \multiVTrandom have the highest proportion of cases for which they did not predict $\mathbf{{v}}$, but a shared FN frame or a co-hyponym.  
    Moreover, \mamTrandom falls short against {both} VL models  across the board on FN frames, in contrast to our findings above on \gls{map} on verb \textit{types}. 
    E.g.,~\mamVTrandom predicted \word{fry} instead of reference \word{roast}, but both evoke the Apply\_heat frame and are co-hyponyms of \word{cook}, while 
     \mamTrandom predicted \word{put} (Placing; \word{move}; see also Fig.~\ref{fig:verb_generalization_analysis} of App.~\ref{app:mam}). 
    Noteworthy is also that \mamVTrandom is best on frame-level generalization \textit{with} visual input during inference, while  \multiVTrandom is the 
    best model on the co-hyponymy 
    relation \textit{without} visual input. 
    The patterns indicate that, first, \textbf{the visual modality is beneficial for frame semantics,}  
    and second, \textbf{\gls{vl} action learning fosters event knowledge, while \gls{vl} action--effect learning fosters lexico-taxonomic knowledge.}

\paragraph{Qualitative Analysis.}
Through introspection of individual verb classes, we found the \textbf{visually perceivable effect to be beneficial for the majority} of our examined classes. 
    For example, the accuracy of \mamVTrandom on Seen \word{whip} and \word{bake}  drops significantly (\mbox{$-24$pp} and \mbox{$-20.7$pp}, resp.), 
    when it is tested with language input only; examples for Unseen are \word{crumple},  and \word{wring} (\mbox{$-9.1$pp}, \mbox{$-6.3$pp}, resp.). 
    However, \textbf{for several classes the linguistic context alone gives a strong cue}. 
    \word{Manicure}, e.g.,~often co-occurs with the word \word{nail}, and \word{sharpen} with \word{knife} 
    (we refer to App.~\ref{app:mam}, Fig.~\ref{fig:verb_pred} for an example). 
For an example of extrapolations of visual effect similarities for capturing behavior equivalence, see Appendix~\ref{app:mam}, Fig.~\ref{fig:verb_generalization}. 

\begin{table}[t]
    \centering
    \begin{tabular}{@{~}l@{~}|@{~}l@{~}|@{~}r@{~}}
    \toprule 
    Test Set Input  & Model  & Accuracy  \\
    \midrule
    \multirow{3}{*}{Multimodal} 
    & Random baseline  & 0.27  \\
    & \multiVTrandom   & 59.2  \\
    & \memVTrandom   & \textbf{59.9} \\
     \midrule
    \multirow{3}{*}{Video} 
    & \multiVTrandom   & 58.6  \\
    & \memVTrandom  & 57.2  \\
    & \memVrandom  & \textbf{59.7}  \\
    \bottomrule
    \end{tabular}
    \caption{Results (\%) for MEP on 
    (1) Multimodal (video clips + subtitles) and (2) Video (video clips) test input.
    \label{tab:final_mep_results}}
\end{table}

\begin{table*}[t]
\small
	\centering
	\begin{tabular}{@{~}l@{~}|@{~}rrrrr@{~}r@{~}|@{~}r@{~}||@{~}rrrr@{~}}
		\toprule
		Model & Stack & Roll  & Grasp& Break  & \underline{Slide} & Bounce& Macro  
            & 1 & 2 & 3 & 4\\ 
		\midrule
            \mamT & 30.0 & 20.0 & 24.0 & 27.0 & 24.0 & 23.0 & 24.7 
                & 22.5 & 22.5 & 22.3 & 22.3  \\
            \mamVT & 22.0  & 25.0 & 33.0 & 26.0 & 30.0 & 23.0 & 26.5 
                     & 25.0  & 25.0 & 25.2 & 25.1 \\
            \multiVT & \textbf{34.0}$^\ast$ & 24.0 & 30.0 & 26.0 & \textbf{40.0}$^\ast$ & \textbf{38.0}$^\ast$ & \textbf{32.0}$^\ast$
                   & 27.4 & 27.4 & 27.6 & 27.7 \\
            \flava & 19.0 & 26.9 & \textbf{34.5} & 25.5 & 23.8 & 24.0 & 25.5 & 25.8 & 21.5 & 21.0 & 21.7 \\
            \roberta & 27.4 & {24.9} & 22.8 & \textbf{31.0}$^\ast$ & 26.6 & 25.5 & 26.4 
                 & 45.8 & 19.4 & 15.1 & 15.0 \\
            \avglm$^\dagger$ & 29.1 & \textbf{29.0} & 28.4 & 30.7 & 20.9 & 20.4 & 26.3  
                 & 31.4 & 25.2 & 8.5 & 38.6 \\
  		\bottomrule
	\end{tabular}
\caption{\protect \gls{prost}: \textbf{Left}: Results (accuracy \%) 
	  across six affordances.  
   \textbf{Right}: Position accuracy across the correct answer's position. The more balanced, the more robust. 
	 $^\dagger$Results taken from \gls{prost} \cite{aroca-ouellette-etal-2021-prost}. Statistically significant differences ($p < 0.05$)\footnote{We use student’s t-test \cite{Fisher1949-qa} 
	 	between the top performing model per affordance group and the other models.} are indicated with $^\ast$.
   }\label{tab:prost_main_result}
\end{table*}

\begin{table*}[ht]
\small
	\centering
	\begin{tabular}{@{~}l@{~}|rrrrrr|r@{~}}
		\toprule
		Model & Stack $\downarrow$  & Roll $\downarrow$ & Grasp $\downarrow$ & Break $\downarrow$ & \underline{Slide} $\downarrow$ & Bounce $\downarrow$ & Macro Average $\downarrow$ \\ 
		\midrule
            \mamT  & 60.0 & 20.0 & 32.0 & 14.0 & 40.0 & 34.0 & 33.3 \\
             \mamVT  & 4.0 & 30.0 & 62.0 & \textbf{12.0} & 20.0 & 6.0 & 22.3 \\
            \multiVT & 68.0 & 32.0 & 56.0 & \textbf{12.0} & 80.0 & 76.0 & 54.0 \\
        \flava & 4.5  & 45.2 & 68.6 & 41.9 & 39.6 & 31.7 & 38.6 \\
         \roberta & \textbf{0.1}  & \textbf{14.5} & \textbf{5.2} & 51.9 & 17.5 & \textbf{2.4} & 15.3 \\
        \avglm & 11.4 & 14.7 & 12.7 & 17.8 & \textbf{12.8} & 9.6 & \textbf{13.2} \\  
  	\bottomrule
	\end{tabular}
\caption{\protect \gls{prost}: Absolute difference in accuracy between the original template and its inverse across six affordances. The lower the better the model's true question understanding.} \label{tab:prost_template_result}
\end{table*}

\subsection{Intrinsic Tasks:  \gls{mep}}
\label{subsect:mep}

    As shown in Table~\ref{tab:final_mep_results}, the accuracy of the multimodal and vision-only model, \memVTrandom and \memVrandom, is comparable on visual effect inference on  their respective input modes ($59.9$\% and $59.7$\%). 
    This indicates that linguistic knowledge priors do not contribute much to visual action--effect comprehension. 
    Comparing these models with a variant whose language encoder is initialized with \roberta before multimodal pretraining supports this---it yields a neglectable difference in accuracy   ($60$\% and $59.6$\% on \gls{vl} and V test input, resp., no table shown). 
    On the other hand, on video-only inference,   \multiVT is slightly more effective than  \memVTrandom  ($+1.4$pp).
    Thus, a linguistic encoder that is trained jointly on both, visual effect and \textit{linguistic} action prediction (\multiVT) may benefit   
    \textit{video-only} effect inference. 
    We refer to Figures~\ref{fig:mem_entity_equivalence} and~\ref{fig:mem_video_incorrect} for examples of successful entity equivalence reasoning, and failures due to ambiguous reference. 

\subsection{Probing Task: \gls{prost}}
\label{sec:extrinsic_results}

\paragraph{Pretraining models on both, 
\textit{visual} action \textit{and}  effect prediction in the video domain, is 
beneficial for learning affordance knowledge.}

\label{subsect:affordance_results}

 As Table \ref{tab:prost_main_result} (left) shows,  \multiVT, which is additionally trained to observe the action process \textit{and} its perceptual effect on objects
 obtains the best Macro Average~($32$\%), with a difference to \avglm of up to $+19.1$pp (\prostafford{slide}). 
 Notably, it achieves better results on average compared to \flava ($+16.2$pp). This indicates that modeling visual action--effects in the temporal dimension plays a crucial role in effectively encoding latent affordance concepts.
 Note that \prostafford{slide} is the only affordance that is contained in our \gls{cae} training data, yet,  
even on \prostafford{slide}, 
 \mamVT 
 yields a higher accuracy ($+6$pp, Tab.~\ref{tab:prost_main_result}) than 
 its variant trained only on textual training items (subtitles), \mamT. 

 
 Beneficial is moreover pretraining towards visual effect prediction (i.e.,~\gls{mem}), as the comparison of \multiVT against \mamVT shows ($+5.5$pp on average). 
 The former significantly outperforms all other models also on two affordances that are not covered by \gls{cae}'s train split (\prostafford{stack} ($34$\%)  and \prostafford{bounce} ($38$\%)).
That is, \multiVT learns to extrapolate to unseen actions. \\
Our models are at least comparable to the pure LMs on all individual affordances except for \prostafford{break} (\roberta, $31$\%). When compared to \flava, our models perform slightly worse in the case of \prostafford{grasp}, though the performance difference is not significant.
Introspection showed  the performance discrepancy between the affordances is not due to seen vs. unseen,  \textit{objects}, 
in fact, 
all 
are seen (cf.~Tab.~\ref{tab:prost_action_object},  App.~\ref{app:prost}). 
However, we found that the errors made by our model in \prostafford{break} and \prostafford{grasp} are related to the frequency of objects in the \gls{cae} training set. In other words, our models tend to pick the most common seen object among the $4$ choices as its prediction, e.g., wrongly selecting \prostobject{sugar} as the graspable object.
\paragraph{Model Robustness.}
 We test the models for robustness to the order of answer choices  and to template inversion (difference in accuracy on affordance/non-affordance), allowing deeper insights into the models' proper affordance knowledge and language understanding ability, respectively.  
 Table~\ref{tab:prost_main_result} (right) shows that all our models display the most balanced effectiveness across different answer positions, indicating robustness against syntactic change of the answer position, i.e., the order in which the correct (non-)affordable object is presented in the context. In contrast, text-based models like \roberta and \avglm are significantly affected, while \flava slightly favors the first answer position.

In terms of robustness to template inversion,  as shown in Table~\ref{tab:prost_template_result}, \multiVT, the overall most effective model in accuracy, is the least balanced, it has on average a $54$pp higher accuracy on {inverses}. 
Our second overall best model, \mamVT, is by far more robust (avg.~$22.3$pp diff.) and is
 better than \flava. 
This result is somewhat surprising since \flava is expected to have a stronger linguistic understanding due to its additional pretraining on textual corpora. \textbf{The pure language models turn out to be most robust to inverses} 
(avg.~$13.2$pp and $15.3$pp for \avglm and \roberta, resp.), indicating a higher ability of  \textit{true language} understanding.\footnote{Recall our \gls{cae} pretraining data has noisy subtitles including automatically transcribed speech, 
which may hinder language understanding.} See Appendix~\ref{app:prost}, Table~\ref{tab:prost_template_result_breakdown} for the breakdown of the individual results on original/inversed templates.

\section{Discussion}
\label{sec:discussions}
The intrinsic results suggest that both modalities contribute beneficially towards our goal of encoding behavior and entity equivalence. 
Crucially, our results on the \gls{prost} affordance probing task strongly support our joint action--effect modeling in the video domain: \multiVT is the most effective model.
It is also most robust in terms of language-only (esp.~\gls{prost}) and visual-only input (\gls{mep}) inference modes.
We hypothesize that \multiVT is more encouraged in taking both modalities into account due to its training objectives on both, vision-conditioned \textit{linguistic} action prediction (\gls{mam}) and language-conditioned \textit{visual} effect prediction (\gls{mem}). 

\section{Conclusions}
\label{sec:conclusions}
We explore the acquisition of
affordance properties--behavior~and entity equivalence through large-scale causal action--effect pretraining in the video domain. 
To this end, we
augment an existing instructional video dataset and introduce two pretraining tasks, \gls{mam} \& \gls{mem}. 
Our empirical results show the successful incorporation of these foundamental properties of affordance.
Furthermore, the joint-task-trained model outperforms   linguistic counterparts on an 
affordance probing task.
Future work would benefit from a cleaner dataset to explicitly examine the contribution of action/effect modeling, and
%
further exploration of multi-task training. 

\section{Limitations}
\label{sec:limitations}
In this work, we seek to validate the induction of two affordance properties through large-scale self-supervised visual--linguistic pretraining. It is important to acknowledge that affordance knowledge encompasses a range of concepts \cite{DBLP:conf/cvpr/Tool_ZhuZZ15, DBLP:journals/corr/PartAfford_Xu2022}, and our assessment is currently limited to a linguistic probing task. Future work is necessary to investigate whether the representation derived from causal action--effect modeling can effectively address other affordance downstream tasks, e.g.,~purpose-driven affordance understanding \cite{DBLP:conf/ijcai/LuoZ00T21, DBLP:journals/ijcv/ZhaiLZCT22}.
Furthermore, it is worth investigating the potential applications of the encoded latent affordance representations in areas like procedural tasks \cite{DBLP:conf/cvpr/ZhouMKSN23} and robot learning \cite{DBLP:conf/cvpr/BahlMCJP23}.
When considering the use of the video domain for large-scale pretraining, one has to remark on certain shortcomings compared to a controllable embodied environment. Firstly, the temporal misalignment of the video clips and subtitles poses a challenge for cross-modal learning \cite{DBLP:conf/cvpr/CrossTask_ZhukovACFLS19,DBLP:conf/cvpr/MiechASLSZ20}. Moreover, many subtitles consist of incomplete sentences, as most of them are automatically transcribed based on an ongoing speech within fixed time intervals \cite{DBLP:journals/corr/KuehneYT19}, restricting the linguistic capabilities of models trained on such data. 
Secondly, it is computationally infeasible to ensure quality control on the web-crawled video data, particularly when it comes to establishing clear temporal boundaries and localization between the pre-condition, action process, and the resulting effect. 
Additionally, eliminating background objects proves to be a challenging task in this context.
Lastly, to draw more conclusive insights into the intrinsic evaluations, it would be beneficial to have a human-annotated test set. 
However, this approach can be prohibitively costly, necessitating further research into semi-automatic methods for conducting such annotations \cite{DBLP:conf/cvpr/HuangBDGFN18, DBLP:conf/eccv/ActionFormerZhangWL22, DBLP:journals/corr/StepFormer_Dvornik22}.
Regarding our methodology, it is important to note that our focus is on presenting proof-of-concept pretraining tasks rather than achieving optimal performance. Hence, we have not conducted an extensive hyperparameter search, and the results reported in this study are obtained from a single seed run.

\section*{Acknowledgments}We are grateful to Matteo Bortoletto for proofreading and providing valuable suggestions for the paper, as well as to Pavel Denisov for offering technical assistance. Lastly, we thank the anonymous reviews for their constructive and valuable feedback.

\bibliography{anthology, refs_affordances}

\begin{thebibliography}{69}
\expandafter\ifx\csname natexlab\endcsname\relax\def\natexlab#1{#1}\fi

\bibitem[{Ard{\'o}n et~al.(2021)Ard{\'o}n, Pairet, Lohan, Ramamoorthy, and Petrick}]{Ardon2021-os}
Paola Ard{\'o}n, {\`E}ric Pairet, Katrin~S Lohan, Subramanian Ramamoorthy, and Ronald P~A Petrick. 2021.
\newblock \href {http://arxiv.org/abs/2105.06706} {Building affordance relations for robotic agents - a review}.

\bibitem[{Aroca-Ouellette et~al.(2021)Aroca-Ouellette, Paik, Roncone, and Kann}]{aroca-ouellette-etal-2021-prost}
St{\'e}phane Aroca-Ouellette, Cory Paik, Alessandro Roncone, and Katharina Kann. 2021.
\newblock \href {https://doi.org/10.18653/v1/2021.findings-acl.404} {{PROST}: {P}hysical reasoning about objects through space and time}.
\newblock In \emph{Findings of the Association for Computational Linguistics: ACL-IJCNLP 2021}, pages 4597--4608, Online. Association for Computational Linguistics.

\bibitem[{Bahl et~al.(2023)Bahl, Mendonca, Chen, Jain, and Pathak}]{DBLP:conf/cvpr/BahlMCJP23}
Shikhar Bahl, Russell Mendonca, Lili Chen, Unnat Jain, and Deepak Pathak. 2023.
\newblock \href {https://doi.org/10.1109/CVPR52729.2023.01324} {Affordances from human videos as a versatile representation for robotics}.
\newblock In \emph{{IEEE/CVF} Conference on Computer Vision and Pattern Recognition, {CVPR} 2023, Vancouver, BC, Canada, June 17-24, 2023}, pages 1--13. {IEEE}.

\bibitem[{Baker et~al.(1998)Baker, Fillmore, and Lowe}]{baker-etal-1998-berkeley-framenet}
Collin~F. Baker, Charles~J. Fillmore, and John~B. Lowe. 1998.
\newblock \href {https://doi.org/10.3115/980845.980860} {The {B}erkeley {F}rame{N}et project}.
\newblock In \emph{36th Annual Meeting of the Association for Computational Linguistics and 17th International Conference on Computational Linguistics, Volume 1}, pages 86--90, Montreal, Quebec, Canada. Association for Computational Linguistics.

\bibitem[{Bird and Loper(2004)}]{bird-loper-2004-nltk}
Steven Bird and Edward Loper. 2004.
\newblock \href {https://aclanthology.org/P04-3031} {{NLTK}: The natural language toolkit}.
\newblock In \emph{Proceedings of the {ACL} Interactive Poster and Demonstration Sessions}, pages 214--217, Barcelona, Spain. Association for Computational Linguistics.

\bibitem[{Bisk et~al.(2020{\natexlab{a}})Bisk, Holtzman, Thomason, Andreas, Bengio, Chai, Lapata, Lazaridou, May, Nisnevich, Pinto, and Turian}]{bisk-etal-2020-experience}
Yonatan Bisk, Ari Holtzman, Jesse Thomason, Jacob Andreas, Yoshua Bengio, Joyce Chai, Mirella Lapata, Angeliki Lazaridou, Jonathan May, Aleksandr Nisnevich, Nicolas Pinto, and Joseph Turian. 2020{\natexlab{a}}.
\newblock \href {https://doi.org/10.18653/v1/2020.emnlp-main.703} {Experience grounds language}.
\newblock In \emph{Proceedings of the 2020 Conference on Empirical Methods in Natural Language Processing (EMNLP)}, pages 8718--8735, Online. Association for Computational Linguistics.

\bibitem[{Bisk et~al.(2020{\natexlab{b}})Bisk, Zellers, Bras, Gao, and Choi}]{DBLP:conf/aaai/PIQA_Bisk}
Yonatan Bisk, Rowan Zellers, Ronan~Le Bras, Jianfeng Gao, and Yejin Choi. 2020{\natexlab{b}}.
\newblock \href {https://ojs.aaai.org/index.php/AAAI/article/view/6239} {{PIQA:} reasoning about physical commonsense in natural language}.
\newblock In \emph{The Thirty-Fourth {AAAI} Conference on Artificial Intelligence, {AAAI} 2020, The Thirty-Second Innovative Applications of Artificial Intelligence Conference, {IAAI} 2020, The Tenth {AAAI} Symposium on Educational Advances in Artificial Intelligence, {EAAI} 2020, New York, NY, USA, February 7-12, 2020}, pages 7432--7439. {AAAI} Press.

\bibitem[{Bosselut et~al.(2017)Bosselut, Levy, Holtzman, Ennis, Fox, and Choi}]{Bosselut2017-aj}
Antoine Bosselut, Omer Levy, Ari Holtzman, Corin Ennis, Dieter Fox, and Yejin Choi. 2017.
\newblock \href {http://arxiv.org/abs/1711.05313} {Simulating action dynamics with neural process networks}.

\bibitem[{Brysbaert et~al.(2014)Brysbaert, Warriner, and Kuperman}]{brysbaert_2014}
Marc Brysbaert, Amy~Beth Warriner, and Victor Kuperman. 2014.
\newblock \href {https://doi.org/10.3758/s13428-013-0403-5} {Concreteness ratings for 40 thousand generally known english word lemmas}.
\newblock \emph{Behavior Research Methods}.

\bibitem[{Chao et~al.(2015)Chao, Wang, Mihalcea, and Deng}]{Chao2015-kh}
Yu-Wei Chao, Zhan Wang, Rada Mihalcea, and Jia Deng. 2015.
\newblock \href {http://dx.doi.org/10.1109/CVPR.2015.7299054} {Mining semantic affordances of visual object categories}.
\newblock In \emph{2015 {IEEE} Conference on Computer Vision and Pattern Recognition ({CVPR})}, pages 4259--4267.

\bibitem[{Da\~{g} et~al.(2010)Da\~{g}, Atil, Kalkan, and {\c S}ahin}]{Dag2010-hd}
Nilgun Da\~{g}, Ilkay Atil, Sinan Kalkan, and Erol {\c S}ahin. 2010.
\newblock \href {http://dx.doi.org/10.1109/ICPR.2010.1146} {Learning affordances for categorizing objects and their properties}.
\newblock In \emph{2010 20th International Conference on Pattern Recognition}, pages 3089--3092.

\bibitem[{Dagan et~al.(2023)Dagan, Keller, and Lascarides}]{dagan-etal-2023-learning}
Gautier Dagan, Frank Keller, and Alex Lascarides. 2023.
\newblock \href {https://aclanthology.org/2023.findings-eacl.10} {Learning the effects of physical actions in a multi-modal environment}.
\newblock In \emph{Findings of the Association for Computational Linguistics: EACL 2023}, pages 133--148, Dubrovnik, Croatia. Association for Computational Linguistics.

\bibitem[{Dehban et~al.(2016)Dehban, Jamone, Kampff, and Santos-Victor}]{Dehban2016-ok}
Atabak Dehban, Lorenzo Jamone, Adam~R Kampff, and Jose Santos-Victor. 2016.
\newblock \href {http://dx.doi.org/10.1109/icra.2016.7487691} {Denoising auto-encoders for learning of objects and tools affordances in continuous space}.

\bibitem[{Deng et~al.(2009)Deng, Dong, Socher, Li, Li, and Fei-Fei}]{Deng2009-ku}
Jia Deng, Wei Dong, Richard Socher, Li-Jia Li, Kai Li, and Li~Fei-Fei. 2009.
\newblock \href {http://dx.doi.org/10.1109/CVPR.2009.5206848} {{ImageNet}: A large-scale hierarchical image database}.
\newblock In \emph{2009 {IEEE} Conference on Computer Vision and Pattern Recognition}, pages 248--255.

\bibitem[{Dvornik et~al.(2023)Dvornik, Hadji, Zhang, Derpanis, Garg, Wildes, and Jepson}]{DBLP:journals/corr/StepFormer_Dvornik22}
Nikita Dvornik, Isma Hadji, Ran Zhang, Konstantinos~G. Derpanis, Animesh Garg, Richard~P. Wildes, and Allan~D. Jepson. 2023.
\newblock \href {https://doi.org/10.48550/arXiv.2304.13265} {Stepformer: Self-supervised step discovery and localization in instructional videos}.
\newblock \emph{CoRR}, abs/2304.13265.

\bibitem[{Ebert et~al.(2023)Ebert, Sun, and Pavlick}]{DBLP:journals/corr/Ebert_verb_trajectory}
Dylan Ebert, Chen Sun, and Ellie Pavlick. 2023.
\newblock \href {https://doi.org/10.48550/arXiv.2303.12737} {Comparing trajectory and vision modalities for verb representation}.
\newblock \emph{CoRR}, abs/2303.12737.

\bibitem[{Erk(2007)}]{Erk2007-wl}
Katrin Erk. 2007.
\newblock \href {https://aclanthology.org/P07-1028} {A simple, similarity-based model for selectional preferences}.
\newblock In \emph{Proceedings of the 45th Annual Meeting of the Association of Computational Linguistics}, pages 216--223.

\bibitem[{Feichtenhofer et~al.(2019)Feichtenhofer, Fan, Malik, and He}]{Feichtenhofer2019-il}
Christoph Feichtenhofer, Haoqi Fan, Jitendra Malik, and Kaiming He. 2019.
\newblock \href {http://openaccess.thecvf.com/content_ICCV_2019/html/Feichtenhofer_SlowFast_Networks_for_Video_Recognition_ICCV_2019_paper.html} {Slowfast networks for video recognition}.
\newblock In \emph{Proceedings of the {IEEE/CVF} international conference on computer vision}, pages 6202--6211.

\bibitem[{Fisher(1949)}]{Fisher1949-qa}
Ronald~A Fisher. 1949.
\newblock \href {https://psycnet.apa.org/record/1950-03922-000} {The design of experiments}.

\bibitem[{Gao et~al.(2016)Gao, Doering, Yang, and Chai}]{Gao2016-tj}
Qiaozi Gao, Malcolm Doering, Shaohua Yang, and Joyce Chai. 2016.
\newblock \href {https://www.aclweb.org/anthology/P16-1171.pdf} {Physical causality of action verbs in grounded language understanding}.
\newblock In \emph{Proceedings of the 54th Annual Meeting of the Association for Computational Linguistics (Volume 1: Long Papers)}, pages 1814--1824.

\bibitem[{Gao et~al.(2018)Gao, Yang, Chai, and Vanderwende}]{Gao2018-kc}
Qiaozi Gao, Shaohua Yang, Joyce Chai, and Lucy Vanderwende. 2018.
\newblock \href {https://www.aclweb.org/anthology/P18-1086.pdf} {What action causes this? towards naive physical action-effect prediction}.
\newblock In \emph{Proceedings of the 56th Annual Meeting of the Association for Computational Linguistics (Volume 1: Long Papers)}, pages 934--945.

\bibitem[{Gibson(1977)}]{Gibson1977-rq}
James~J Gibson. 1977.
\newblock \href {https://books.google.com/books?hl=en&lr=&id=b9WWAwAAQBAJ&oi=fnd&pg=PA56&dq=The+theory+of+affordances&ots=KW_qwMioAd&sig=oKzmmRoBsOUNTqtTK3iFQ3Rwz1w} {The theory of affordances}.
\newblock \emph{Hilldale, USA}, 1(2):67--82.

\bibitem[{Hanna et~al.(2022)Hanna, Pedeni, Suglia, Testoni, and Bernardi}]{hanna-etal-2022-act}
Michael Hanna, Federico Pedeni, Alessandro Suglia, Alberto Testoni, and Raffaella Bernardi. 2022.
\newblock \href {https://aclanthology.org/2022.coling-1.495} {{ACT}-thor: A controlled benchmark for embodied action understanding in simulated environments}.
\newblock In \emph{Proceedings of the 29th International Conference on Computational Linguistics}, pages 5597--5612, Gyeongju, Republic of Korea. International Committee on Computational Linguistics.

\bibitem[{Hassanin et~al.(2022)Hassanin, Khan, and Tahtali}]{DBLP:journals/csur/HassaninKT21}
Mohammed Hassanin, Salman~H. Khan, and Murat Tahtali. 2022.
\newblock \href {https://doi.org/10.1145/3446370} {Visual affordance and function understanding: {A} survey}.
\newblock \emph{{ACM} Comput. Surv.}, 54(3):47:1--47:35.

\bibitem[{{He} et~al.(){He}, {Zhang}, {Ren}, and {Sun}}]{He_undated-dx}
{He}, {Zhang}, {Ren}, and {Sun}.
\newblock \href {http://openaccess.thecvf.com/content_cvpr_2016/html/He_Deep_Residual_Learning_CVPR_2016_paper.html} {Deep residual learning for image recognition}.
\newblock \emph{and pattern recognition}.

\bibitem[{He et~al.(2016)He, Zhang, Ren, and Sun}]{DBLP:conf/cvpr/HeZRS16}
Kaiming He, Xiangyu Zhang, Shaoqing Ren, and Jian Sun. 2016.
\newblock \href {https://doi.org/10.1109/CVPR.2016.90} {Deep residual learning for image recognition}.
\newblock In \emph{2016 {IEEE} Conference on Computer Vision and Pattern Recognition, {CVPR} 2016, Las Vegas, NV, USA, June 27-30, 2016}, pages 770--778. {IEEE} Computer Society.

\bibitem[{Huang et~al.(2018)Huang, Buch, Dery, Garg, Fei{-}Fei, and Niebles}]{DBLP:conf/cvpr/HuangBDGFN18}
De{-}An Huang, Shyamal Buch, Lucio~M. Dery, Animesh Garg, Li~Fei{-}Fei, and Juan~Carlos Niebles. 2018.
\newblock \href {https://doi.org/10.1109/CVPR.2018.00623} {Finding "it": Weakly-supervised reference-aware visual grounding in instructional videos}.
\newblock In \emph{2018 {IEEE} Conference on Computer Vision and Pattern Recognition, {CVPR} 2018, Salt Lake City, UT, USA, June 18-22, 2018}, pages 5948--5957. Computer Vision Foundation / {IEEE} Computer Society.

\bibitem[{Huang et~al.(2022)Huang, Xia, Xiao, Chan, Liang, Florence, Zeng, Tompson, Mordatch, Chebotar, Sermanet, Jackson, Brown, Luu, Levine, Hausman, and Ichter}]{DBLP:conf/corl/Inner_mono_Huang22}
Wenlong Huang, Fei Xia, Ted Xiao, Harris Chan, Jacky Liang, Pete Florence, Andy Zeng, Jonathan Tompson, Igor Mordatch, Yevgen Chebotar, Pierre Sermanet, Tomas Jackson, Noah Brown, Linda Luu, Sergey Levine, Karol Hausman, and Brian Ichter. 2022.
\newblock \href {https://proceedings.mlr.press/v205/huang23c.html} {Inner monologue: Embodied reasoning through planning with language models}.
\newblock In \emph{Conference on Robot Learning, CoRL 2022, 14-18 December 2022, Auckland, New Zealand}, volume 205 of \emph{Proceedings of Machine Learning Research}, pages 1769--1782. {PMLR}.

\bibitem[{Ichter et~al.(2022)Ichter, Brohan, Chebotar, Finn, Hausman, Herzog, Ho, Ibarz, Irpan, Jang, Julian, Kalashnikov, Levine, Lu, Parada, Rao, Sermanet, Toshev, Vanhoucke, Xia, Xiao, Xu, Yan, Brown, Ahn, Cortes, Sievers, Tan, Xu, Reyes, Rettinghouse, Quiambao, Pastor, Luu, Lee, Kuang, Jesmonth, Joshi, Jeffrey, Ruano, Hsu, Gopalakrishnan, David, Zeng, and Fu}]{DBLP:conf/corl/SayCan_Ichter22}
Brian Ichter, Anthony Brohan, Yevgen Chebotar, Chelsea Finn, Karol Hausman, Alexander Herzog, Daniel Ho, Julian Ibarz, Alex Irpan, Eric Jang, Ryan Julian, Dmitry Kalashnikov, Sergey Levine, Yao Lu, Carolina Parada, Kanishka Rao, Pierre Sermanet, Alexander Toshev, Vincent Vanhoucke, Fei Xia, Ted Xiao, Peng Xu, Mengyuan Yan, Noah Brown, Michael Ahn, Omar Cortes, Nicolas Sievers, Clayton Tan, Sichun Xu, Diego Reyes, Jarek Rettinghouse, Jornell Quiambao, Peter Pastor, Linda Luu, Kuang{-}Huei Lee, Yuheng Kuang, Sally Jesmonth, Nikhil~J. Joshi, Kyle Jeffrey, Rosario~Jauregui Ruano, Jasmine Hsu, Keerthana Gopalakrishnan, Byron David, Andy Zeng, and Chuyuan~Kelly Fu. 2022.
\newblock \href {https://proceedings.mlr.press/v205/ichter23a.html} {Do as {I} can, not as {I} say: Grounding language in robotic affordances}.
\newblock In \emph{Conference on Robot Learning, CoRL 2022, 14-18 December 2022, Auckland, New Zealand}, volume 205 of \emph{Proceedings of Machine Learning Research}, pages 287--318. {PMLR}.

\bibitem[{Jaramillo-Cabrera et~al.(2019)Jaramillo-Cabrera, Morales, and Martinez-Carranza}]{Jaramillo-Cabrera2019-oh}
Esteban Jaramillo-Cabrera, Eduardo~F Morales, and Jose Martinez-Carranza. 2019.
\newblock \href {https://doi.org/10.1177/1059712319839057} {Enhancing object, action, and effect recognition using probabilistic affordances}.
\newblock \emph{Adapt. Behav.}, 27(5):295--306.

\bibitem[{J{\'{o}}zefowicz et~al.(2016)J{\'{o}}zefowicz, Vinyals, Schuster, Shazeer, and Wu}]{NCEsoftmax}
Rafal J{\'{o}}zefowicz, Oriol Vinyals, Mike Schuster, Noam Shazeer, and Yonghui Wu. 2016.
\newblock \href {http://arxiv.org/abs/1602.02410} {Exploring the limits of language modeling}.
\newblock \emph{CoRR}, abs/1602.02410.

\bibitem[{Kay et~al.(2017)Kay, Carreira, Simonyan, Zhang, Hillier, Vijayanarasimhan, Viola, Green, Back, Natsev, Suleyman, and Zisserman}]{Kay2017-zh}
Will Kay, Joao Carreira, Karen Simonyan, Brian Zhang, Chloe Hillier, Sudheendra Vijayanarasimhan, Fabio Viola, Tim Green, Trevor Back, Paul Natsev, Mustafa Suleyman, and Andrew Zisserman. 2017.
\newblock \href {http://arxiv.org/abs/1705.06950} {The kinetics human action video dataset}.

\bibitem[{Kipper et~al.(2006)Kipper, Korhonen, Ryant, and Palmer}]{kipper-etal-2006-extending}
Karin Kipper, Anna Korhonen, Neville Ryant, and Martha Palmer. 2006.
\newblock \href {http://www.lrec-conf.org/proceedings/lrec2006/pdf/468_pdf.pdf} {Extending {V}erb{N}et with novel verb classes}.
\newblock In \emph{Proceedings of the Fifth International Conference on Language Resources and Evaluation ({LREC}{'}06)}, Genoa, Italy. European Language Resources Association (ELRA).

\bibitem[{Kolve et~al.(2017)Kolve, Mottaghi, Gordon, Zhu, Gupta, and Farhadi}]{AI2-THOR_Kolve}
Eric Kolve, Roozbeh Mottaghi, Daniel Gordon, Yuke Zhu, Abhinav Gupta, and Ali Farhadi. 2017.
\newblock \href {http://arxiv.org/abs/1712.05474} {{AI2-THOR:} an interactive 3d environment for visual {AI}}.
\newblock \emph{CoRR}, abs/1712.05474.

\bibitem[{Kuehne et~al.(2019{\natexlab{a}})Kuehne, Iqbal, Richard, and Gall}]{YT_Kuehne2019}
Hilde Kuehne, Ahsan Iqbal, Alexander Richard, and Juergen Gall. 2019{\natexlab{a}}.
\newblock \href {http://arxiv.org/abs/1906.01012} {Mining youtube - {A} dataset for learning fine-grained action concepts from webly supervised video data}.
\newblock \emph{CoRR}, abs/1906.01012.

\bibitem[{Kuehne et~al.(2019{\natexlab{b}})Kuehne, Iqbal, Richard, and Gall}]{DBLP:journals/corr/KuehneYT19}
Hilde Kuehne, Ahsan Iqbal, Alexander Richard, and Juergen Gall. 2019{\natexlab{b}}.
\newblock \href {http://arxiv.org/abs/1906.01012} {Mining youtube - {A} dataset for learning fine-grained action concepts from webly supervised video data}.
\newblock \emph{CoRR}, abs/1906.01012.

\bibitem[{Levin and Malka~Rappaport(2010)}]{Levin_undated-pc}
Beth Levin and Hovav Malka~Rappaport. 2010.
\newblock Lexicalized scales and verbs of scalar change.

\bibitem[{Li et~al.(2020)Li, Chen, Cheng, Gan, Yu, and Liu}]{li-etal-2020-hero}
Linjie Li, Yen-Chun Chen, Yu~Cheng, Zhe Gan, Licheng Yu, and Jingjing Liu. 2020.
\newblock \href {https://doi.org/10.18653/v1/2020.emnlp-main.161} {{HERO}: Hierarchical encoder for {V}ideo+{L}anguage omni-representation pre-training}.
\newblock In \emph{Proceedings of the 2020 Conference on Empirical Methods in Natural Language Processing (EMNLP)}, pages 2046--2065, Online. Association for Computational Linguistics.

\bibitem[{Loshchilov and Hutter(2019)}]{adamW}
Ilya Loshchilov and Frank Hutter. 2019.
\newblock \href {https://openreview.net/forum?id=Bkg6RiCqY7} {Decoupled weight decay regularization}.
\newblock In \emph{7th International Conference on Learning Representations, {ICLR} 2019, New Orleans, LA, USA, May 6-9, 2019}. OpenReview.net.

\bibitem[{Loureiro and Jorge(2018)}]{loureiro-jorge-2018-affordance}
Daniel Loureiro and Al{\'\i}pio Jorge. 2018.
\newblock \href {https://doi.org/10.18653/v1/W18-5514} {Affordance extraction and inference based on semantic role labeling}.
\newblock In \emph{Proceedings of the First Workshop on Fact Extraction and {VER}ification ({FEVER})}, pages 91--96, Brussels, Belgium. Association for Computational Linguistics.

\bibitem[{Lu et~al.(2022)Lu, Zhai, Luo, Kang, and Cao}]{LuPAD-L}
Liangsheng Lu, Wei Zhai, Hongchen Luo, Yu~Kang, and Yang Cao. 2022.
\newblock \href {http://arxiv.org/abs/2202.12076} {Phrase-based affordance detection via cyclic bilateral interaction}.
\newblock \emph{CoRR}, abs/2202.12076.

\bibitem[{Luo et~al.(2021)Luo, Zhai, Zhang, Cao, and Tao}]{DBLP:conf/ijcai/LuoZ00T21}
Hongchen Luo, Wei Zhai, Jing Zhang, Yang Cao, and Dacheng Tao. 2021.
\newblock \href {https://doi.org/10.24963/ijcai.2021/124} {One-shot affordance detection}.
\newblock In \emph{Proceedings of the Thirtieth International Joint Conference on Artificial Intelligence, {IJCAI} 2021, Virtual Event / Montreal, Canada, 19-27 August 2021}, pages 895--901. ijcai.org.

\bibitem[{Merullo et~al.(2022)Merullo, Ebert, Eickhoff, and Pavlick}]{merullo-etal-2022-pretraining}
Jack Merullo, Dylan Ebert, Carsten Eickhoff, and Ellie Pavlick. 2022.
\newblock \href {https://doi.org/10.18653/v1/2022.starsem-1.23} {Pretraining on interactions for learning grounded affordance representations}.
\newblock In \emph{Proceedings of the 11th Joint Conference on Lexical and Computational Semantics}, pages 258--277, Seattle, Washington. Association for Computational Linguistics.

\bibitem[{Miech et~al.(2020)Miech, Alayrac, Smaira, Laptev, Sivic, and Zisserman}]{DBLP:conf/cvpr/MiechASLSZ20}
Antoine Miech, Jean{-}Baptiste Alayrac, Lucas Smaira, Ivan Laptev, Josef Sivic, and Andrew Zisserman. 2020.
\newblock \href {https://doi.org/10.1109/CVPR42600.2020.00990} {End-to-end learning of visual representations from uncurated instructional videos}.
\newblock In \emph{2020 {IEEE/CVF} Conference on Computer Vision and Pattern Recognition, {CVPR} 2020, Seattle, WA, USA, June 13-19, 2020}, pages 9876--9886. Computer Vision Foundation / {IEEE}.

\bibitem[{Miech et~al.(2019)Miech, Zhukov, Alayrac, Tapaswi, Laptev, and Sivic}]{DBLP:conf/iccv/MiechZATLS19}
Antoine Miech, Dimitri Zhukov, Jean{-}Baptiste Alayrac, Makarand Tapaswi, Ivan Laptev, and Josef Sivic. 2019.
\newblock \href {https://doi.org/10.1109/ICCV.2019.00272} {Howto100m: Learning a text-video embedding by watching hundred million narrated video clips}.
\newblock In \emph{2019 {IEEE/CVF} International Conference on Computer Vision, {ICCV} 2019, Seoul, Korea (South), October 27 - November 2, 2019}, pages 2630--2640. {IEEE}.

\bibitem[{Miller(1994)}]{miller-1994-wordnet}
George~A. Miller. 1994.
\newblock \href {https://aclanthology.org/H94-1111} {{W}ord{N}et: A lexical database for {E}nglish}.
\newblock In \emph{{H}uman {L}anguage {T}echnology: Proceedings of a Workshop held at {P}lainsboro, {N}ew {J}ersey, {M}arch 8-11, 1994}.

\bibitem[{Nagarajan and Grauman(2020)}]{DBLP:conf/nips/NagarajanG20}
Tushar Nagarajan and Kristen Grauman. 2020.
\newblock \href {https://proceedings.neurips.cc/paper/2020/hash/15825aee15eb335cc13f9b559f166ee8-Abstract.html} {Learning affordance landscapes for interaction exploration in 3d environments}.
\newblock In \emph{Advances in Neural Information Processing Systems 33: Annual Conference on Neural Information Processing Systems 2020, NeurIPS 2020, December 6-12, 2020, virtual}.

\bibitem[{Ott et~al.(2018)Ott, Edunov, Grangier, and Auli}]{DBLP:conf/wmt/OttEGA18}
Myle Ott, Sergey Edunov, David Grangier, and Michael Auli. 2018.
\newblock \href {https://doi.org/10.18653/v1/w18-6301} {Scaling neural machine translation}.
\newblock In \emph{Proceedings of the Third Conference on Machine Translation: Research Papers, {WMT} 2018, Belgium, Brussels, October 31 - November 1, 2018}, pages 1--9. Association for Computational Linguistics.

\bibitem[{Paik et~al.(2021)Paik, Aroca-Ouellette, Roncone, and Kann}]{paik-etal-2021-world}
Cory Paik, St{\'e}phane Aroca-Ouellette, Alessandro Roncone, and Katharina Kann. 2021.
\newblock \href {https://doi.org/10.18653/v1/2021.emnlp-main.63} {{T}he {W}orld of an {O}ctopus: {H}ow {R}eporting {B}ias {I}nfluences a {L}anguage {M}odel{'}s {P}erception of {C}olor}.
\newblock In \emph{Proceedings of the 2021 Conference on Empirical Methods in Natural Language Processing}, pages 823--835, Online and Punta Cana, Dominican Republic. Association for Computational Linguistics.

\bibitem[{Pantel et~al.(2007)Pantel, Bhagat, Coppola, Chklovski, and Hovy}]{Pantel2007-ci}
Patrick Pantel, Rahul Bhagat, Bonaventura Coppola, Timothy Chklovski, and Eduard Hovy. 2007.
\newblock \href {https://www.aclweb.org/anthology/N07-1071.pdf} {{ISP}: Learning inferential selectional preferences}.
\newblock In \emph{Human Language Technologies 2007: The Conference of the North American Chapter of the Association for Computational Linguistics; Proceedings of the Main Conference}, pages 564--571.

\bibitem[{Persiani and Hellstr{\"o}m(2019)}]{Persiani2019-hd}
Michele Persiani and Thomas Hellstr{\"o}m. 2019.
\newblock \href {https://www.diva-portal.org/smash/record.jsf?pid=diva2:1351658} {Unsupervised inference of object affordance from text corpora}.
\newblock In \emph{22nd Nordic Conference on Computational Linguistics ({NoDaLiDa'19)}, September 30 -- October 2, 2019, Turku, Finland}. Association for Computational Linguistics.

\bibitem[{Petroni et~al.(2019)Petroni, Rockt{\"a}schel, Riedel, Lewis, Bakhtin, Wu, and Miller}]{petroni-etal-2019-language}
Fabio Petroni, Tim Rockt{\"a}schel, Sebastian Riedel, Patrick Lewis, Anton Bakhtin, Yuxiang Wu, and Alexander Miller. 2019.
\newblock \href {https://doi.org/10.18653/v1/D19-1250} {Language models as knowledge bases?}
\newblock In \emph{Proceedings of the 2019 Conference on Empirical Methods in Natural Language Processing and the 9th International Joint Conference on Natural Language Processing (EMNLP-IJCNLP)}, pages 2463--2473, Hong Kong, China. Association for Computational Linguistics.

\bibitem[{Regneri et~al.(2013)Regneri, Rohrbach, Wetzel, Thater, Schiele, and Pinkal}]{regneri-etal-2013-grounding}
Michaela Regneri, Marcus Rohrbach, Dominikus Wetzel, Stefan Thater, Bernt Schiele, and Manfred Pinkal. 2013.
\newblock \href {https://doi.org/10.1162/tacl_a_00207} {Grounding action descriptions in videos}.
\newblock \emph{Transactions of the Association for Computational Linguistics}, 1:25--36.

\bibitem[{{\c S}ahin et~al.(2007){\c S}ahin, {\c C}akmak, Do{\u g}ar, U{\u g}ur, and {\"U}{\c c}oluk}]{Sahin2007-pq}
Erol {\c S}ahin, Maya {\c C}akmak, Mehmet~R Do{\u g}ar, Emre U{\u g}ur, and G{\"o}kt{\"u}rk {\"U}{\c c}oluk. 2007.
\newblock \href {https://doi.org/10.1177/1059712307084689} {To afford or not to afford: A new formalization of affordances toward {Affordance-Based} robot control}.
\newblock \emph{Adapt. Behav.}, 15(4):447--472.

\bibitem[{Sampat et~al.(2022{\natexlab{a}})Sampat, Banerjee, Yang, and Baral}]{sampat-etal-2022-learning}
Shailaja~Keyur Sampat, Pratyay Banerjee, Yezhou Yang, and Chitta Baral. 2022{\natexlab{a}}.
\newblock \href {https://aclanthology.org/2022.findings-emnlp.436} {Learning action-effect dynamics for hypothetical vision-language reasoning task}.
\newblock In \emph{Findings of the Association for Computational Linguistics: EMNLP 2022}, pages 5914--5924, Abu Dhabi, United Arab Emirates. Association for Computational Linguistics.

\bibitem[{Sampat et~al.(2022{\natexlab{b}})Sampat, Patel, Das, Yang, and Baral}]{RACsurvey}
Shailaja~Keyur Sampat, Maitreya Patel, Subhasish Das, Yezhou Yang, and Chitta Baral. 2022{\natexlab{b}}.
\newblock \href {https://doi.org/10.48550/arXiv.2207.07568} {Reasoning about actions over visual and linguistic modalities: {A} survey}.
\newblock \emph{CoRR}, abs/2207.07568.

\bibitem[{Shwartz and Choi(2020)}]{shwartz-choi-2020-neural}
Vered Shwartz and Yejin Choi. 2020.
\newblock \href {https://doi.org/10.18653/v1/2020.coling-main.605} {Do neural language models overcome reporting bias?}
\newblock In \emph{Proceedings of the 28th International Conference on Computational Linguistics}, pages 6863--6870, Barcelona, Spain (Online). International Committee on Computational Linguistics.

\bibitem[{Singh et~al.(2022)Singh, Hu, Goswami, Couairon, Galuba, Rohrbach, and Kiela}]{DBLP:conf/cvpr/FLAVA22}
Amanpreet Singh, Ronghang Hu, Vedanuj Goswami, Guillaume Couairon, Wojciech Galuba, Marcus Rohrbach, and Douwe Kiela. 2022.
\newblock \href {https://doi.org/10.1109/CVPR52688.2022.01519} {{FLAVA:} {A} foundational language and vision alignment model}.
\newblock In \emph{{IEEE/CVF} Conference on Computer Vision and Pattern Recognition, {CVPR} 2022, New Orleans, LA, USA, June 18-24, 2022}, pages 15617--15629. {IEEE}.

\bibitem[{Tang et~al.(2019)Tang, Ding, Rao, Zheng, Zhang, Zhao, Lu, and Zhou}]{DBLP:conf/cvpr/COIN_TangDRZZZL019}
Yansong Tang, Dajun Ding, Yongming Rao, Yu~Zheng, Danyang Zhang, Lili Zhao, Jiwen Lu, and Jie Zhou. 2019.
\newblock \href {https://doi.org/10.1109/CVPR.2019.00130} {{COIN:} {A} large-scale dataset for comprehensive instructional video analysis}.
\newblock In \emph{{IEEE} Conference on Computer Vision and Pattern Recognition, {CVPR} 2019, Long Beach, CA, USA, June 16-20, 2019}, pages 1207--1216. Computer Vision Foundation / {IEEE}.

\bibitem[{Wolf et~al.(2020)Wolf, Debut, Sanh, Chaumond, Delangue, Moi, Cistac, Rault, Louf, Funtowicz, Davison, Shleifer, von Platen, Ma, Jernite, Plu, Xu, Le~Scao, Gugger, Drame, Lhoest, and Rush}]{wolf-etal-2020-transformers}
Thomas Wolf, Lysandre Debut, Victor Sanh, Julien Chaumond, Clement Delangue, Anthony Moi, Pierric Cistac, Tim Rault, Remi Louf, Morgan Funtowicz, Joe Davison, Sam Shleifer, Patrick von Platen, Clara Ma, Yacine Jernite, Julien Plu, Canwen Xu, Teven Le~Scao, Sylvain Gugger, Mariama Drame, Quentin Lhoest, and Alexander Rush. 2020.
\newblock \href {https://doi.org/10.18653/v1/2020.emnlp-demos.6} {Transformers: State-of-the-art natural language processing}.
\newblock In \emph{Proceedings of the 2020 Conference on Empirical Methods in Natural Language Processing: System Demonstrations}, pages 38--45, Online. Association for Computational Linguistics.

\bibitem[{Xian et~al.(2017)Xian, Schiele, and Akata}]{Xian2017-oi}
Yongqin Xian, Bernt Schiele, and Zeynep Akata. 2017.
\newblock \href {http://openaccess.thecvf.com/content_cvpr_2017/html/Xian_Zero-Shot_Learning_-_CVPR_2017_paper.html} {Zero-shot learning-the good, the bad and the ugly}.
\newblock In \emph{Proceedings of the {IEEE} Conference on Computer Vision and Pattern Recognition}, pages 4582--4591.

\bibitem[{Xu et~al.(2022)Xu, Chen, Wang, Zhu, Zhu, and Huang}]{DBLP:journals/corr/PartAfford_Xu2022}
Chao Xu, Yixin Chen, He~Wang, Song{-}Chun Zhu, Yixin Zhu, and Siyuan Huang. 2022.
\newblock \href {http://arxiv.org/abs/2202.13519} {Partafford: Part-level affordance discovery from 3d objects}.
\newblock \emph{CoRR}, abs/2202.13519.

\bibitem[{Yatskar et~al.(2016)Yatskar, Zettlemoyer, and Farhadi}]{DBLP:conf/cvpr/YatskarZF16}
Mark Yatskar, Luke Zettlemoyer, and Ali Farhadi. 2016.
\newblock \href {https://doi.org/10.1109/CVPR.2016.597} {Situation recognition: Visual semantic role labeling for image understanding}.
\newblock In \emph{2016 {IEEE} Conference on Computer Vision and Pattern Recognition, {CVPR} 2016, Las Vegas, NV, USA, June 27-30, 2016}, pages 5534--5542. {IEEE} Computer Society.

\bibitem[{Zellers et~al.(2021)Zellers, Holtzman, Peters, Mottaghi, Kembhavi, Farhadi, and Choi}]{zellers-etal-2021-piglet}
Rowan Zellers, Ari Holtzman, Matthew Peters, Roozbeh Mottaghi, Aniruddha Kembhavi, Ali Farhadi, and Yejin Choi. 2021.
\newblock \href {https://doi.org/10.18653/v1/2021.acl-long.159} {{PIGL}e{T}: Language grounding through neuro-symbolic interaction in a 3{D} world}.
\newblock In \emph{Proceedings of the 59th Annual Meeting of the Association for Computational Linguistics and the 11th International Joint Conference on Natural Language Processing (Volume 1: Long Papers)}, pages 2040--2050, Online. Association for Computational Linguistics.

\bibitem[{Zhai et~al.(2022)Zhai, Luo, Zhang, Cao, and Tao}]{DBLP:journals/ijcv/ZhaiLZCT22}
Wei Zhai, Hongchen Luo, Jing Zhang, Yang Cao, and Dacheng Tao. 2022.
\newblock \href {https://doi.org/10.1007/s11263-022-01642-4} {One-shot object affordance detection in the wild}.
\newblock \emph{Int. J. Comput. Vis.}, 130(10):2472--2500.

\bibitem[{Zhang et~al.(2022)Zhang, Wu, and Li}]{DBLP:conf/eccv/ActionFormerZhangWL22}
Chen{-}Lin Zhang, Jianxin Wu, and Yin Li. 2022.
\newblock \href {https://doi.org/10.1007/978-3-031-19772-7\_29} {Actionformer: Localizing moments of actions with transformers}.
\newblock In \emph{Computer Vision - {ECCV} 2022 - 17th European Conference, Tel Aviv, Israel, October 23-27, 2022, Proceedings, Part {IV}}, volume 13664 of \emph{Lecture Notes in Computer Science}, pages 492--510. Springer.

\bibitem[{Zhou et~al.(2023)Zhou, Mart{\'{\i}}n{-}Mart{\'{\i}}n, Kapadia, Savarese, and Niebles}]{DBLP:conf/cvpr/ZhouMKSN23}
Honglu Zhou, Roberto Mart{\'{\i}}n{-}Mart{\'{\i}}n, Mubbasir Kapadia, Silvio Savarese, and Juan~Carlos Niebles. 2023.
\newblock \href {https://doi.org/10.1109/CVPR52729.2023.01033} {Procedure-aware pretraining for instructional video understanding}.
\newblock In \emph{{IEEE/CVF} Conference on Computer Vision and Pattern Recognition, {CVPR} 2023, Vancouver, BC, Canada, June 17-24, 2023}, pages 10727--10738. {IEEE}.

\bibitem[{Zhu et~al.(2015)Zhu, Zhao, and Zhu}]{DBLP:conf/cvpr/Tool_ZhuZZ15}
Yixin Zhu, Yibiao Zhao, and Song{-}Chun Zhu. 2015.
\newblock \href {https://doi.org/10.1109/CVPR.2015.7298903} {Understanding tools: Task-oriented object modeling, learning and recognition}.
\newblock In \emph{{IEEE} Conference on Computer Vision and Pattern Recognition, {CVPR} 2015, Boston, MA, USA, June 7-12, 2015}, pages 2855--2864. {IEEE} Computer Society.

\bibitem[{Zhukov et~al.(2019)Zhukov, Alayrac, Cinbis, Fouhey, Laptev, and Sivic}]{DBLP:conf/cvpr/CrossTask_ZhukovACFLS19}
Dimitri Zhukov, Jean{-}Baptiste Alayrac, Ramazan~Gokberk Cinbis, David~F. Fouhey, Ivan Laptev, and Josef Sivic. 2019.
\newblock \href {https://doi.org/10.1109/CVPR.2019.00365} {Cross-task weakly supervised learning from instructional videos}.
\newblock In \emph{{IEEE} Conference on Computer Vision and Pattern Recognition, {CVPR} 2019, Long Beach, CA, USA, June 16-20, 2019}, pages 3537--3545. Computer Vision Foundation / {IEEE}.

\end{thebibliography}
\bibliographystyle{acl_natbib}



\appendix

\section{Appendix}
\label{sec:appendix}
\subsection{Result Verbs: Preprocessing Details}\label{app:result_verbs}
\begin{table*}[t]
\begin{tabular}{@{~}llcc@{~}}
\toprule
Judgement &    & Visualness & Effect-causing  \\ \midrule
Sure Result Verbs &  Verb Senses &   & \\
\midrule
attach  &  [attach-22.3-2-1$^\ast$, attach$^\star$] &  \checkmark  &  \checkmark \\
bend  &  [bend-45.2$^\ast$] &  \checkmark  &  \checkmark \\
chop &  [chop-18.2$^\ast$, chop-21.2-2$^\ast$, chop$^\star$] &  \checkmark  &  \checkmark \\
stretch &  [stretch-45.2$^\ast$, stretch$^\star$] &  \checkmark  &  \checkmark \\
tie &  [tie-22.4$^\ast$, tie-22.1-2$^\ast$, tie$^star$] &  \checkmark  &  \checkmark \\
\midrule
Unsure Result Verbs &  Verb Senses &   & \\ \midrule
activate  &  [activate-45.4$^\ast$] &  ?  &  \checkmark \\
block &  [block$^\star$] &  \checkmark & ?  \\
carve &  [carve-23.3$^\ast$, carve-21.2-2$^\ast$, carve$^\star$] &  \checkmark & ?  \\
sniff &  [sniff$^\star$] &  \checkmark & ?  \\
warm &  [warm-45.4$^\ast$] &  ? & \checkmark  \\
\bottomrule
\end{tabular}
 \caption{Result verbs: sure \& unsure cases examples. The verb senses are sourced from different resources: $^\ast$ represents VerbNet while $\star$ indicates imSitu (no specific verb sense is recorded). "\checkmark" implies the property is satisfied, whereas "?" indicates uncertainty. \label{tab:sure_unsure_result_verbs}}
\end{table*}
\paragraph{VerbNet.}
We use VerbNet 3.3 listed in this repository: \url{https://github.com/cu-clear/verbnet} and the semantic link to find the mapping between VerbNet and FrameNet: \url{https://github.com/cu-clear/semlink/tree/master/instances}.

\paragraph{imSitu.}
The list of invalid imSitu roles at the second position we excluded is ["place", "tool", "location", "manner", "instrument", "listener", "container", "model", "suspect", "victimpart", "addressee", "confronted", "start", 'message', 'skill', 'ailment', 'focus',
'resource', 'experiencer', 'phenomenon', 'agentpart', 'coagent', 'end', 'recipient', 'audience', 'blow', 'supported', 'interviewee', 'destination', 'source', 'carrier', 'entityhelped', 'center', 'reciever', 'event', 'naggedperson', 'obstacle', 'stake', 'coparticipant', 'seller', 'performer', 'student', 'giver', 'reference', 'adressee', 'competition', 'occasion', 'image', 'coagentpart', 'bodypart', 'boringthing', 'victim', 'follower', 'perceiver', 'imitation', 'admired',
'chasee', 'undergoer', 'path', 'shelter', 'restrained'].
The imSitu dataset \cite{DBLP:conf/cvpr/YatskarZF16} could be downloaded under the link: 
\mbox{\url{https://github.com/my89/imSitu}}. 

\paragraph{FrameNet.}
We request version 1.7 from this link: \url{https://framenet.icsi.berkeley.edu/fndrupal/framenet_request_data}.

\paragraph{Result Verbs: Sure Cases v.s. Unsure Cases.}
Following the series of automatic steps described in Section~\ref{subsec:result_verbs}, we derive in total $\textbf{377}$ result verb candidates, including $\textbf{236}$ sure cases and $\textbf{141}$ unsure cases, along with their corresponding $\textbf{150}$ unique FrameNet frames. In this work, only the sure cases are utilized for locating the video clips.
Some examples of sure cases v.s. unsure cases are shown in Table~\ref{tab:sure_unsure_result_verbs}.

\subsection{HowTo100M Subtitles: Preprocessing Details}\label{app:subtitles}
\begin{figure*}
    \centering
    \resizebox{\textwidth}{!}{\includegraphics{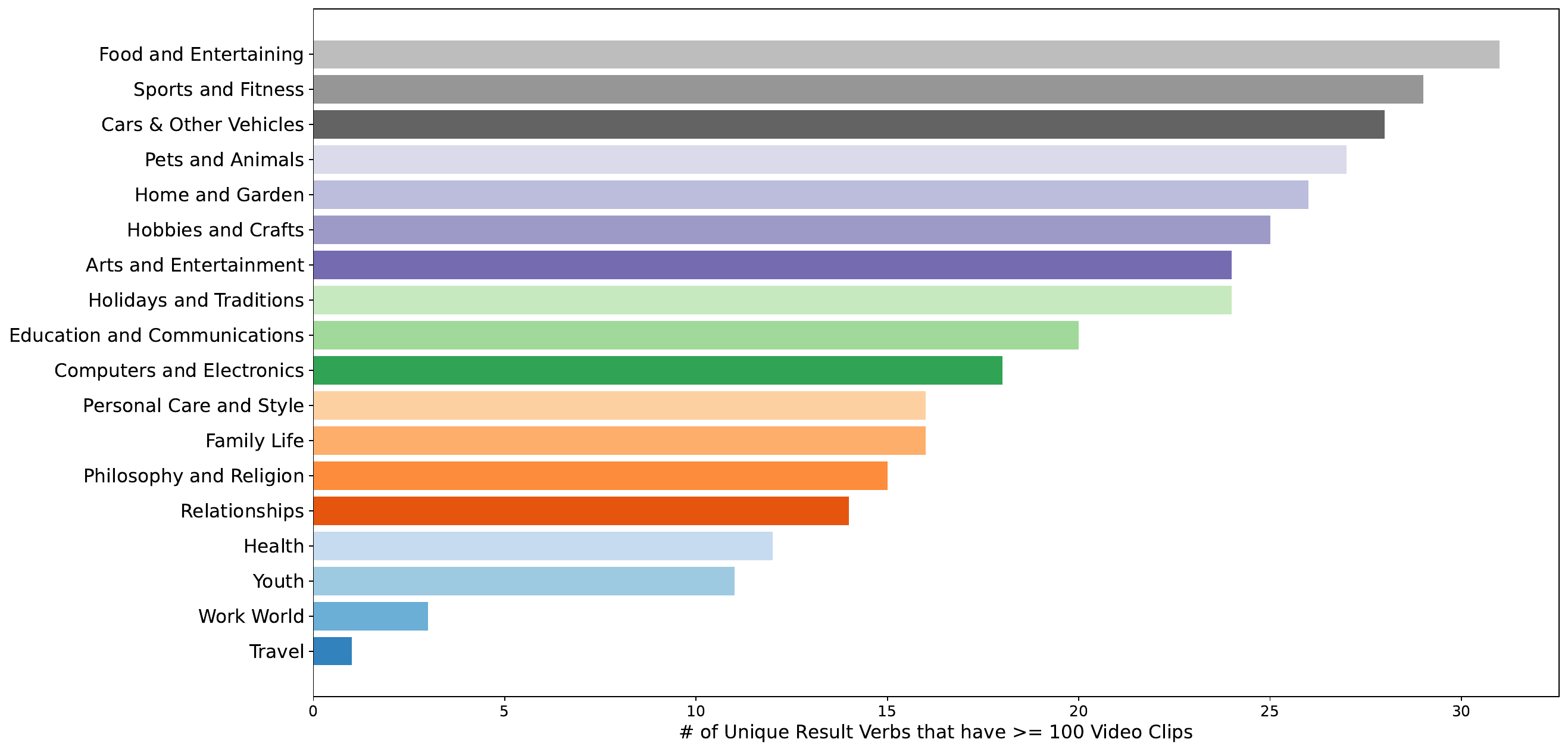}
    }
    \caption{Number of unique result verbs across video categories}
\label{fig:result_verb_type_counts_by_domains}
\end{figure*}

We focus on $13$ video categories (the released file, \texttt{HowTo100M\_v1.csv}, contains $19$ categories) that have a denser distribution of result verbs according to a preliminary study conducted on a subset containing 500 videos from each category. In particular, a video category of our interest should have more than $15$ unique result verb types, and each of the result verb types within should have more than $100$ video clips. The distribution of result verb type across video categories computed during the preliminary study can be seen in Figure~\ref{fig:result_verb_type_counts_by_domains}. 

To further downsample the video pool, we select the top $15$ viewed videos per wikiHow task id (\citet{DBLP:conf/iccv/MiechZATLS19} search and collect videos based on wikiHow task title. On average, there are 48 videos per task id).

We use the pre-processed version of clip-subtitle, \texttt{raw\_caption\_superclean.json}.
All the relevant files can be downloaded from this link: \url{https://www.di.ens.fr/willow/research/howto100m/}.

\begin{figure*}
    \centering
    \resizebox{\textwidth}{!}{\includegraphics{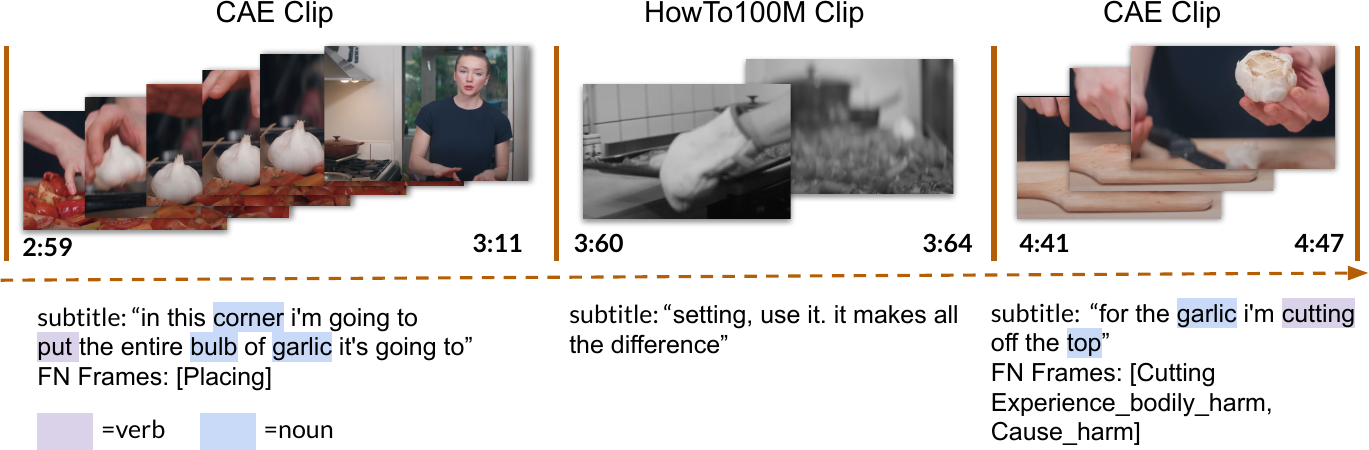}
    }
    \caption{Clip--subtitle pairs: \gls{cae} vs. HowTo100M, \gls{cae} targets action-centric video clips in the HowTo100M}
\label{fig:cae_howto100m}
\end{figure*}

The comparison between extracted \gls{cae} clips  and the non action-centric clip of HowTo100M can be found in Figure~\ref{fig:cae_howto100m}.

\subsection{\gls{cae} Split Statistics}\label{app:split_stats}
\begin{figure*}
    \centering
    \resizebox{\textwidth}{!}{\includegraphics{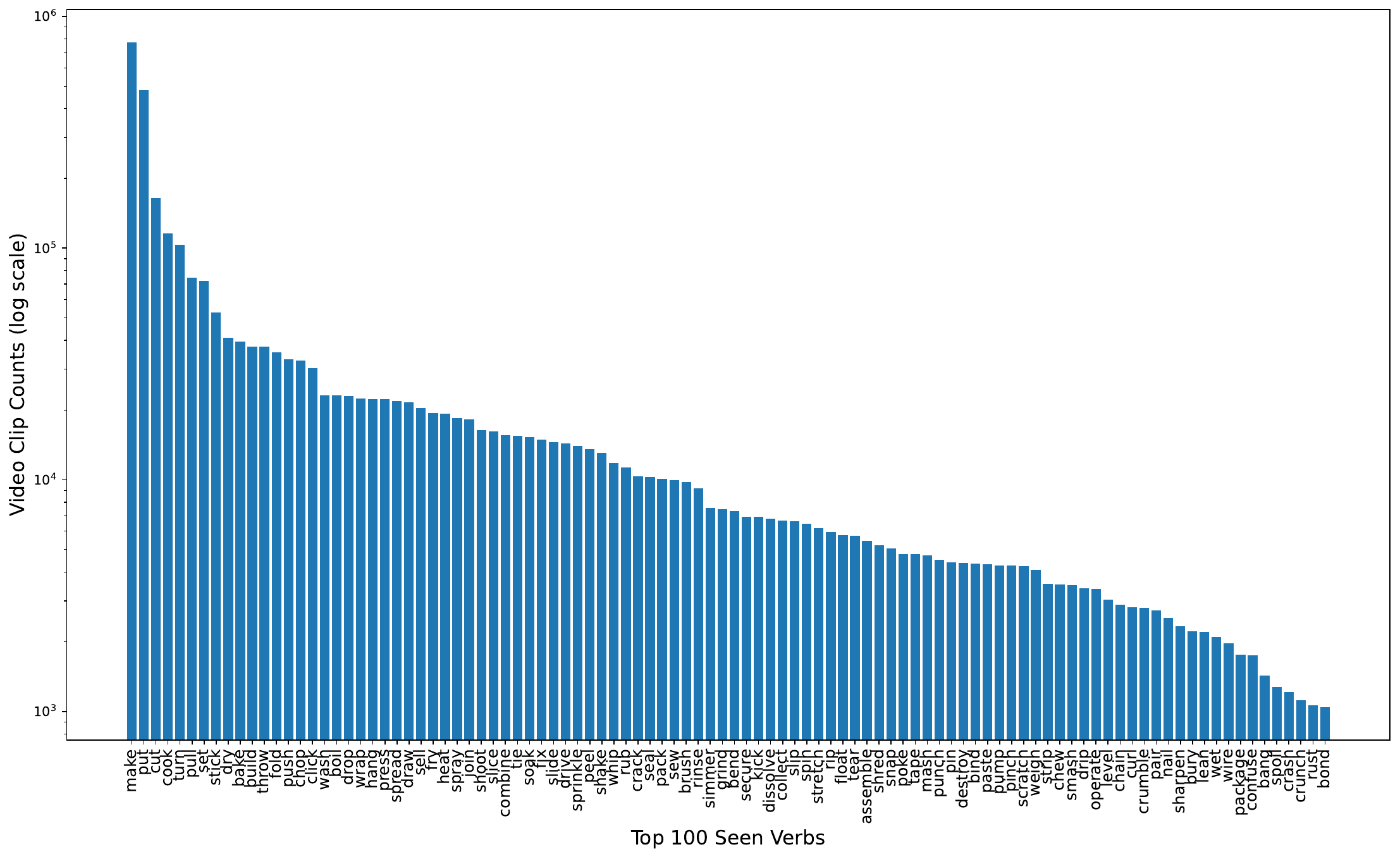}
    }
    \caption{Video clips counts (log scale) of top $100$ seen verb classes in the train set}
\label{fig:top_100_seen_verb_train_instances}
\end{figure*}

\begin{figure*}
    \centering
    \resizebox{\textwidth}{!}{\includegraphics[keepaspectratio]{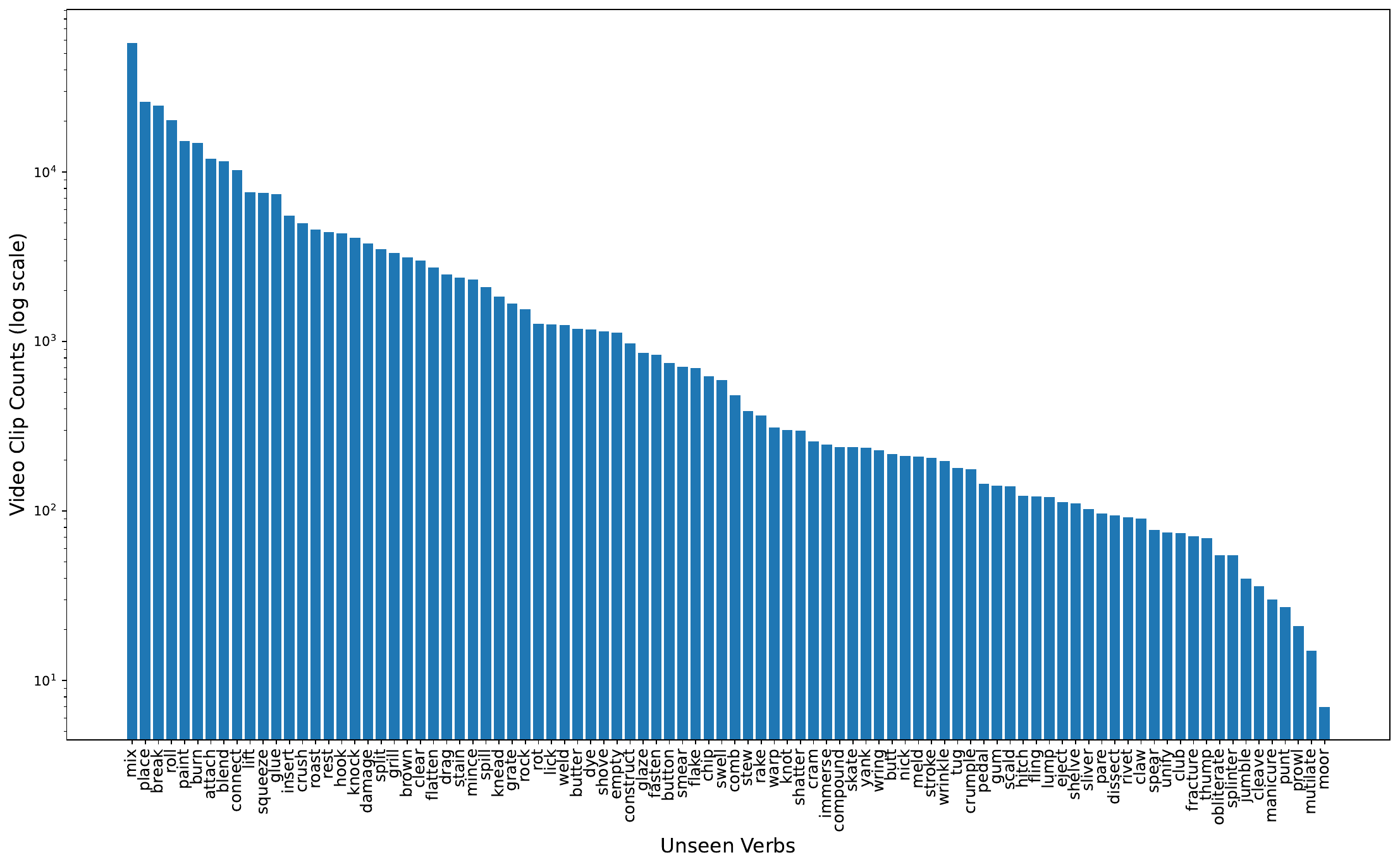}
    }
    \caption{Video clips counts (log scale) of all unseen verb classes in the test set}
\label{fig:unseen_verb_test_instances}
\end{figure*}

\paragraph{Split Details.}
Within a FrameNet frame, we randomly assign (seed 42) 80\%/20\% of the verbs (lexical units) to seen/unseen verb classes. 
As the video frame visual feature extractor is trained with  Kinetics400, we ensure no information leakage by controlling the seen verb classes that do not contain Kinetics400 \cite{Kay2017-zh} verb types.
To control the split, we adopt the recommended practices by \citet{Xian2017-oi}. For seen verb classes, the corresponding video clips ratio of train-dev-test follows 80\%-10\%-10\% whereas for unseen verb classes, it follows 0\%-50\%-50\%. As for the video domain, validation and test set are ensured to be similar. 
\paragraph{Statistics.}
For the top $20$ and bottom $20$ seen verb classes with their top $5$ nouns combinations and the corresponding FrameNet frames they can evoke, refer to Tables~\ref{tab:top_20_seen_verbs} and  \ref{tab:bottom_20_seen_verbs} respectively.
As for the unseen verb classes, refer to Tables~\ref{tab:top_20_unseen_verbs} and  \ref{tab:bottom_20_unseen_verbs}.
The video clip counts (log scale) of top $100$ seen verbs in the train set and unseen verbs ($93$ classes) in the test set can be found in Figures~\ref{fig:top_100_seen_verb_train_instances} and \ref{fig:unseen_verb_test_instances} respectively.

\begin{figure*}
    \centering
    \resizebox{\textwidth}{!}{\includegraphics{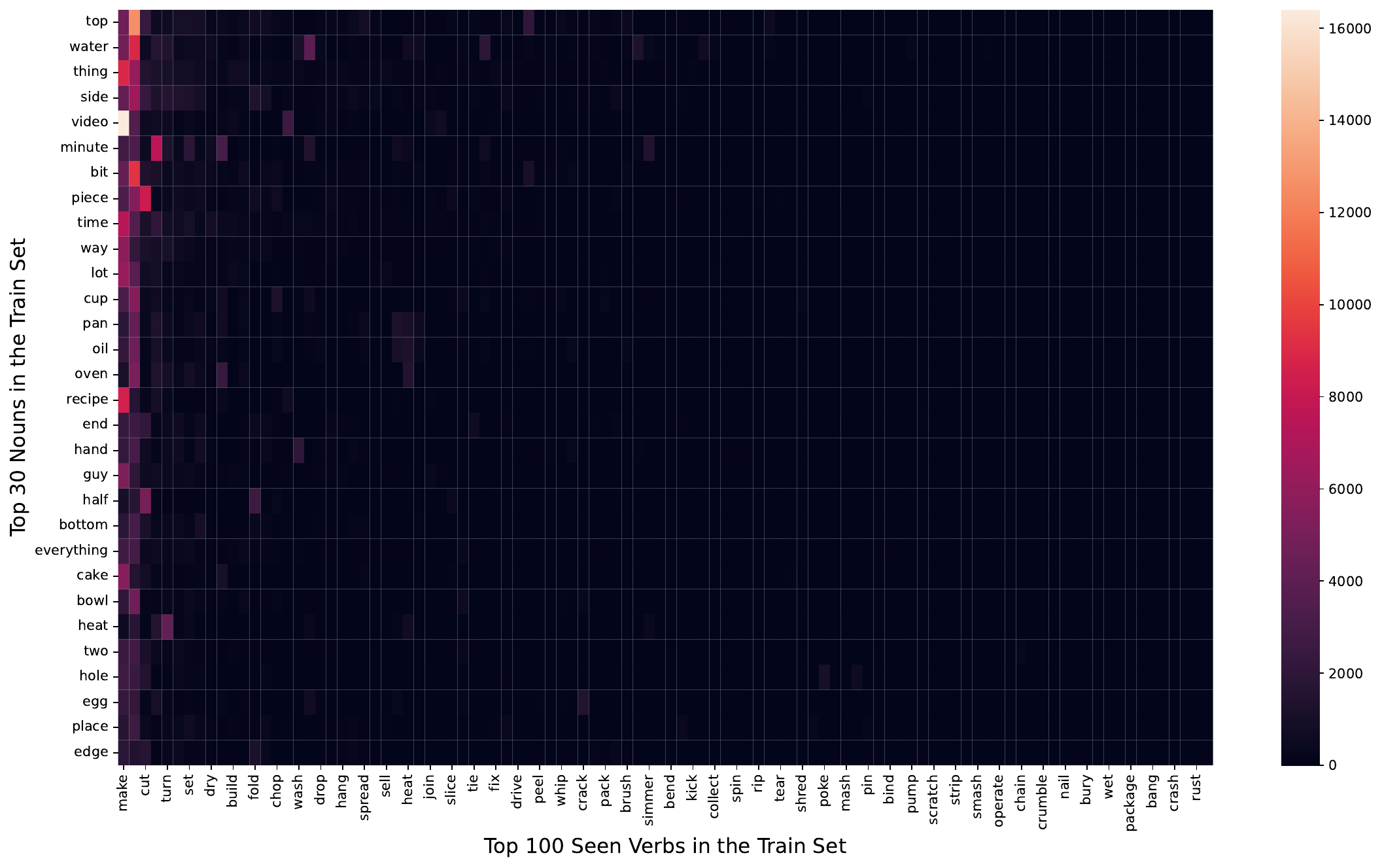}
    }
    \caption{The co-occurrence heatmap between the top $100$ seen verb classes and the top $30$ nouns in the train set}
\label{fig:seen_verb_noun_heatmap_train_set}
\end{figure*}

\begin{figure*}
    \centering
    \resizebox{\textwidth}{!}{\includegraphics{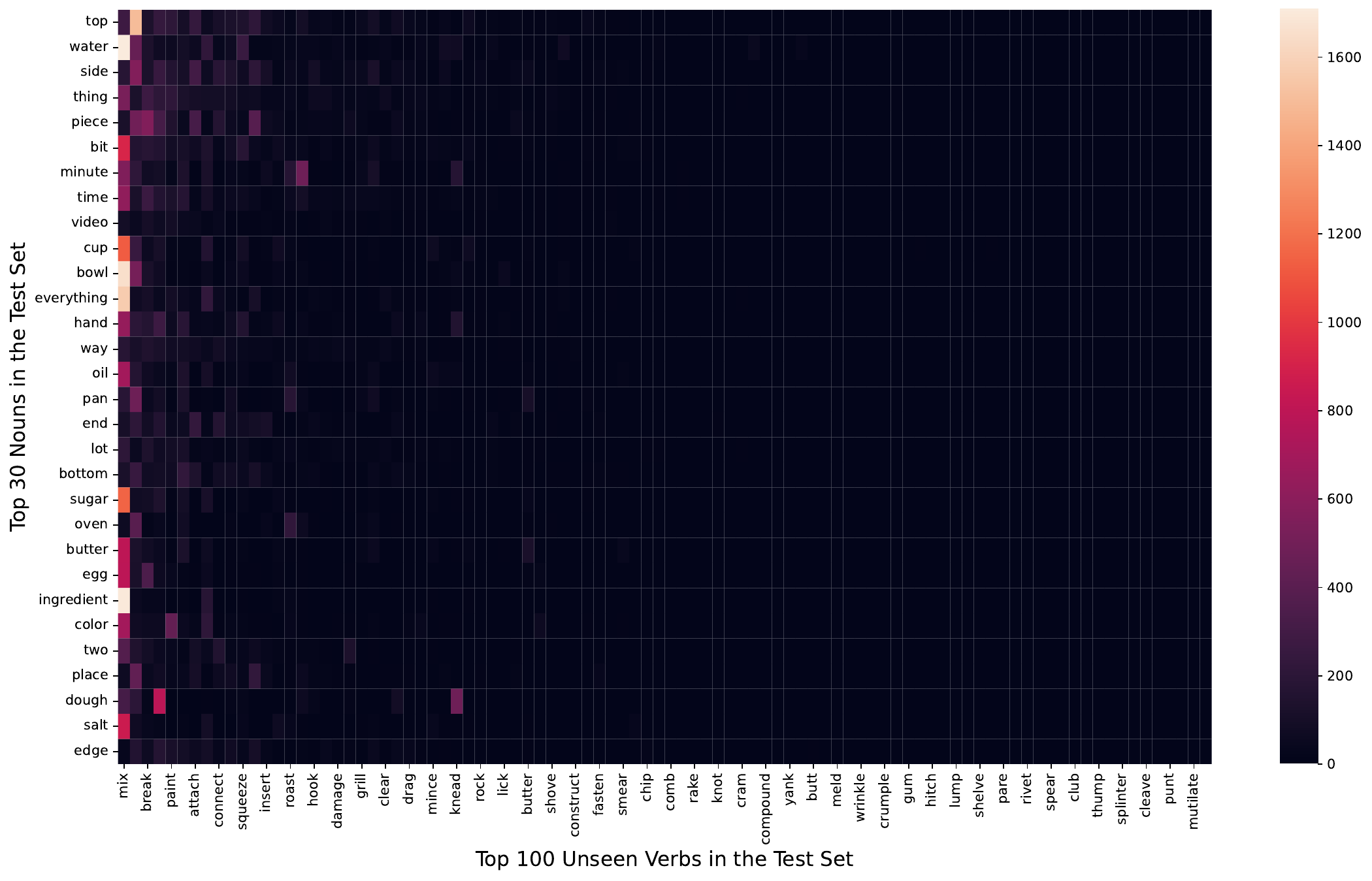}
    }
    \caption{The co-occurrence heatmap between top $100$ unseen verb classes and top $30$ nouns in the test set}
\label{fig:unseen_verb_noun_heatmap_test_set}
\end{figure*}

The co-occurrence heatmap between the top $100$ seen verbs and the top $30$ nouns in the train set is shown in Figure~\ref{fig:seen_verb_noun_heatmap_train_set}. As for that of unseen verbs ($93$ classes) and the top $30$ nouns in the test set is shown in Figure~\ref{fig:unseen_verb_noun_heatmap_test_set}.

\begin{table*}
\resizebox{\linewidth}{!}{
\begin{tabular}{@{~}lll@{~}}
\toprule
Top 20 Seen Verbs &                             Top 5 Nouns &                                     FrameNet frames \\
\midrule
make  &       [video, thing, recipe, time, lot] &                                         [Building] \\
put   &          [top, bit, water, side, thing] &                                          [Placing] \\
cut   &         [piece, half, knife, top, side] &      [Experience\_bodily\_harm, Cause\_harm, Cutting] \\
cook  &       [minute, time, cook, water, heat] &        [Apply\_heat, Cooking\_creation, Absorb\_heat] \\
turn  &      [heat, side, water, minute, light] &                     [Cause\_change, Undergo\_change] \\
pull  &          [side, loop, top, yarn, thing] &  [Cause\_motion, Manipulation, Earnings\_and\_losses] \\
set   &        [minute, side, timer, time, top] &  [Intentionally\_create, Arranging,  \\
      &    & Change\_of\_consistency] \\
stick &          [bottom, side, top, hand, pan] &  [Cause\_motion, Placing, Attaching,  \\
 &           &  Being\_attached] \\
dry   &        [time, hour, minute, dry, water] &                                  [Cause\_to\_be\_dry] \\
bake  &  [soda, powder, minute, oven, teaspoon] &                                 [Cooking\_creation] \\
build &      [thing, house, video, time, build] &                                         [Building] \\
throw &         [thing, bit, water, top, stuff] &                                     [Cause\_motion] \\
fold  &       [half, side, edge, paper, corner] &                                        [Reshaping] \\
push  &         [side, button, top, place, way] &  [Cause\_motion, Manipulation,  \\
 &           &  Cause\_change\_of\_position\_on\_a\_scale] \\
chop  &      [onion, garlic, cup, chop, tomato] &                              [Cause\_harm, Cutting] \\
click &    [video, link, button, channel, icon] &               [Cause\_impact, Impact, Motion\_noise] \\
wash  &         [hand, water, hair, wash, face] &                               [Removing, Grooming] \\
boil  &         [water, minute, boil, egg, cup] &              [Cause\_harm, Apply\_heat, Absorb\_heat] \\
drop  &        [drop, water, comment, oil, top] &                                     [Cause\_motion] \\
wrap  &         [wrap, paper, yarn, wire, tape] &                                          [Placing] \\
\bottomrule
\end{tabular}
}
 \caption{Top $20$ seen verbs and their top $5$ noun combinations \label{tab:top_20_seen_verbs}}
\end{table*}

\begin{table*}
\resizebox{\linewidth}{!}{
\begin{tabular}{@{~}lll@{~}}
\toprule
Bottom 20 Seen Verbs &                                Top 5 Nouns &                 FrameNet frames \\
\midrule
total       &             [cup, dollar, inch, car, gram] &      [Amounting\_to, Adding\_up] \\
decay       &          [matter, decay, wood, time, tree] &                      [Rotting] \\
lash        &       [line, lash, mascara, bottom, thing] &                    [Attaching] \\
belt        &          [belt, pulley, thing, side, drum] &                   [Cause\_harm] \\
cement      &        [place, thing, post, cement, piece] &                    [Attaching] \\
pulverize   &  [processor, thing, blender, garlic, salt] &                     [Grinding] \\
demolish    &  [thing, everything, whole, house, button] &                   [Destroying] \\
paddle      &           [paddle, pool, way, river, time] &          [Corporal\_punishment] \\
collide     &           [way, hold, another, fact, time] &                 [Cause\_impact] \\
plaster     &     [wall, plaster, giordano, video, time] &                    [Attaching] \\
extinguish  &         [fire, flame, time, water, candle] &             [Putting\_out\_fire] \\
boat        &           [boat, water, store, type, hour] &              [Operate\_vehicle] \\
devastate   &          [garden, worm, frost, plant, guy] &                   [Destroying] \\
consolidate &  [item, power, everything, space, surface] &          [Cause\_to\_amalgamate] \\
vaporize    &          [chamomile, water, oil, day, gas] &  [Destroying, Change\_of\_phase] \\
commute     &           [work, people, time, day, thing] &                       [Travel] \\
shush       &           [mouth, kill, shush, cup, sauce] &                    [Silencing] \\
exterminate &          [rat, bedbug, time, garden, pest] &                      [Killing] \\
hurl        &             [time, point, way, wall, wolf] &                 [Cause\_motion] \\
paw         &               [ear, paw, hand, food, luke] &                 [Manipulation] \\
\bottomrule
\end{tabular}
}
 \caption{Bottom 20 seen verbs and their top 5 noun combinations \label{tab:bottom_20_seen_verbs}}
\end{table*}

\begin{table*}
\resizebox{\linewidth}{!}{
\begin{tabular}{@{~}lll@{~}}
\toprule
Top 20 Unseen Verbs &                            Top 5 Nouns &                                     FrameNet frames \\ \midrule
mix     &       [bowl, ingredient, water, everything, mix] &                              [Cause\_to\_amalgamate] \\
place   &                    [top, side, bowl, piece, pan] &                                          [Placing] \\
break   &                 [piece, egg, thing, time, break] &  [Experience\_bodily\_harm, Cause\_harm,  \\
        &               &  Cause\_to\_fragment,
  Render\_nonfunctional] \\
roll    &                  [dough, ball, roll, pin, piece] &  [Cause\_motion, Reshaping, Motion,  \\
        &               &  Mass\_motion, Body\_movement]\\
paint   &                 [paint, color, wall, thing, top] &  [Filling, Create\_physical\_artwork, \\
&                 &   Create\_representation] \\
burn    &                 [burn, bottom, hand, time, fire] &               [Experience\_bodily\_harm, Cause\_harm] \\
attach  &                    [piece, side, top, end, wire] &                                        [Attaching] \\
blend   &  [blender, everything, color, water, ingredient] &                [Amalgamation, Cause\_to\_amalgamate] \\
connect &                   [wire, side, line, piece, end] &                                        [Attaching] \\
lift    &                    [side, top, lid, thing, foot] &                                     [Cause\_motion] \\
squeeze &                 [juice, lemon, water, lime, bit] &                                     [Manipulation] \\
glue    &                  [piece, glue, place, top, side] &                              [Building, Attaching] \\
insert  &          [hook, hole, toothpick, needle, center] &                                          [Placing] \\
crush   &             [garlic, pepper, tomato, ice, clove] &                  [Cause\_harm, Reshaping, Grinding] \\
roast   &               [oven, minute, pan, chicken, seed] &                                       [Apply\_heat] \\
rest    &                  [minute, hour, top, time, oven] &                                          [Placing] \\
hook    &                  [wire, hook, hose, side, thing] &                                        [Attaching] \\
knock   &                  [door, thing, knock, sock, air] &                                     [Cause\_motion] \\
damage  &                  [plant, hair, root, skin, area] &                                         [Damaging] \\
split   &                 [half, two, wood, middle, piece] &                                [Cause\_to\_fragment] \\
\bottomrule
\end{tabular}
}
 \caption{Top 20 unseen verbs and their top 5 noun combinations \label{tab:top_20_unseen_verbs}}
\end{table*}

\begin{table*}
\resizebox{\linewidth}{!}{
\begin{tabular}{@{~}lll@{~}}
\toprule
Bottom 20 Unseen Verbs &                              Top 5 Nouns &                       FrameNet frames \\
\midrule
eject      &          [water, cd, button, air, shell] &                           [Removing] \\
shelve     &        [unit, shelf, side, thing, space] &                            [Placing] \\
sliver     &        [almond, cup, nut, onion, garlic] &                  [Cause\_to\_fragment] \\
pare       &        [knife, side, skin, line, chisel] &                            [Cutting] \\
dissect    &   [earthworm, poem, dissect, image, way] &                  [Cause\_to\_fragment] \\
rivet      &     [rivet, handle, place, tang, bottom] &                          [Attaching] \\
claw       &             [way, claw, hand, crab, top] &                         [Cause\_harm] \\
spear      &          [spear, fish, hole, stick, lot] &                         [Cause\_harm] \\
unify      &        [color, team, look, field, thing] &  [Amalgamation, Cause\_to\_amalgamate] \\
club       &     [club, food, another, friend, final] &                         [Cause\_harm] \\
fracture   &      [bone, fracture, wall, time, crack] &      [Cause\_harm, Cause\_to\_fragment] \\
thump      &       [thump, table, video, side, sound] &   [Cause\_impact, Impact, Make\_noise] \\
obliterate &    [time, deck, chop, piece, everything] &                         [Destroying] \\
splinter   &      [wood, splinter, board, side, edge] &                  [Cause\_to\_fragment] \\
jumble     &    [word, stuff, guy, thing, everything] &                [Cause\_to\_amalgamate] \\
cleave     &       [cleave, bit, pinching, wood, log] &                  [Cause\_to\_fragment] \\
manicure   &     [nail, lawn, hand, manicure, course] &                           [Grooming] \\
punt       &          [puppy, thing, ball, way, punt] &                       [Cause\_motion] \\
prowl      &     [prowl, top, treat, attacker, river] &                        [Self\_motion] \\
mutilate   &  [beak, man, listener, madhubala, money] &                         [Cause\_harm] \\
\bottomrule
\end{tabular}
}
 \caption{Bottom 20 unseen verbs and their top 5 noun combinations \label{tab:bottom_20_unseen_verbs}}
\end{table*}

\subsection{Model Details}\label{app:model}
\paragraph{Subtitle Input.}
Subtitles are tokenized with \texttt{RobertaTokenizerFast} that uses byte-level Byte-Pair-Encoding from the Hugging Face library \cite{wolf-etal-2020-transformers}. 
\paragraph{Video Input: Visual Feature Extraction.}
For each video clip, the temporal context is extended to \textbf{3~seconds} before the starting point and after the endpoint of the original time stamp such as to ensure that the pre-condition and the post-condition are sufficiently captured. 
The visual features are then extracted every 2 seconds (0.5 FPS)  with ResNet \cite{He_undated-dx},  pretrained on ImageNet \cite{Deng2009-ku}, and with SlowFast \cite{Feichtenhofer2019-il}, pretrained on Kinectics-400 \cite{Kay2017-zh} for 2D and 3D features, respectively. Therefore, for a video segment $V = \{\vb{v}_i\}_{i=1}^{|V|}$ ($|V|$ is the number of decoded video frames), the resulting visual representation is the concatenation of 2D (2048) and 3D features (2304) ($V \in \mathbb{R}^{|V|}\times 4352$).

\paragraph{HERO Architecture.}
Designed by \citet{li-etal-2020-hero}, it can be viewed as a single-stream vision--language model that captures contextualized video embeddings in a hierarchical fashion.
There are three major components: the \textbf{Input Embedder}, the \textbf{Cross-Modal Transformer}, and the \textbf{Temporal Transformer}. 
The \textbf{Input Embedder} comprises the Visual and Textual Embedder. 
The final textual representation of a sub-word token is obtained by applying a  normalization (LN) layer on top of the sum of (1) the token embedding (2) the position embedding and (3) the segment type embedding (one segment type). 
As for the visual part, the video frame embedding is obtained by projecting the extracted visual features to the same dimension as the token embedding with a fully-connected (FC) layer. 
Similarly, to obtain the final visual representation, (1) the video frame embedding (2) the position embedding and (3) the segment type embedding comprising three types: \texttt{[BEF]} (to denote the pre-condition), \texttt{[ACT]} (to denote the action process), \texttt{[AFT]} (to denote the post-condition) are summed up and fed through a LN layer. Finally, the multimodal input is the concatenation of the textual and the visual representation. 

To learn the implicit association between subtitle tokens and video frames, the \textbf{Cross-Modal Transformer} is used to perform cross-modal attention on the multimodal input. 
The outputs are then contextualized subtitle embeddings and video frame embeddings.

To further obtain temporal-aware video frame embeddings, the local output of video frame embeddings from the Cross-Modal Transformer are fed into the \textbf{Temporal Transformer} yielding the global embeddings.
The final contextualized video frame embeddings are obtained by adding the local and the global embeddings via a residual connection \cite{DBLP:conf/cvpr/HeZRS16}.

\subsection{\gls{mam}}
\label{app:mam}
    During the development stage 
    performed on $10$\%~of the training data and test data
    , we explored three masking strategies   
(see Table~\ref{tab:mam_masking_strategies}): 
        \begin{itemize}
            \item \textbf{Verb-only}: mask the result verb only.
            \item \textbf{Verb-random-joint}: mask the result verb \& and some random words within one data point in the meantime. In this way, more masked tokens have to be reconstructed; thus, it is more challenging.
            \item \textbf{Verb-random-alter}: mask the result verb or random words alternatively, i.e., 50\% of the training data with only the result verb tokens being masked, and the rest with only random tokens being masked. 
        \end{itemize}
The best action generalization ability (highest accuracy on unseen verbs) was yielded by  \textbf{verb-random-joint} (see Tab.~\ref{tab:dev_map_results}); therefore, we adopt this masking strategy for \gls{mam}.

\begin{table*}[t]
    \centering
    \resizebox{0.85\textwidth}{!}{
    \begin{tabular}{@{~}l|l|l@{~}}
    \toprule 
    MAM Masking Strategies  & Original text & Masking Result  \\
    \midrule
    verb only & Chop the carrot into pieces. & \texttt{[MASK]} the carrot into pieces. \\
            & Chop the garlic into small chunks. & \texttt{[MASK]} the garlic into small chunks. \\ \midrule
    verb random joint & Chop the carrot into pieces. & \texttt{[MASK]} the \texttt{[MASK]} into pieces.  \\
    & Chop the garlic into small chunks. & \texttt{[MASK]} the \texttt{[MASK]} into small \texttt{[MASK]}. \\ \midrule
    verb random alter & Chop the carrot into pieces. & \texttt{[MASK]} the carrot into pieces.  \\
     & Chop the garlic into small chunks. & Chop the garlic into \texttt{[MASK]} chunks.
    \\
    \bottomrule 
    \end{tabular}
    }
    \caption{\gls{mam} masking strategies  \label{tab:mam_masking_strategies}}
\end{table*}

\begin{table*}[t]
    \centering
    \begin{tabular}{@{~}l|r|r|r@{~}}
    \toprule 
    MAM Masking Strategies  & Seen Verb Acc & Unseen Verb Acc & Harmonic Mean  \\
    \midrule
    Verb-only & $25.35$ & $12.93$ &  $17.13$ \\ \midrule
    Verb-random-joint & $28.62$ & $\textbf{18.71}$ &  $\textbf{22.63}$ \\ \midrule
    Verb-random-alter & $\textbf{29.74}$ & $17.09$ &  $21.71$ \\
    \bottomrule 
    \end{tabular}
    \caption{Intrinsic evaluation of \gls{mam} during development on predicting the correct seen / unseen verbs when tested on  $10\%$ test set. \label{tab:dev_map_results}}
\end{table*}

\paragraph{Qualitative Analysis.}
\begin{figure*}
    \centering
    \resizebox{\textwidth}{!}{
    \includegraphics{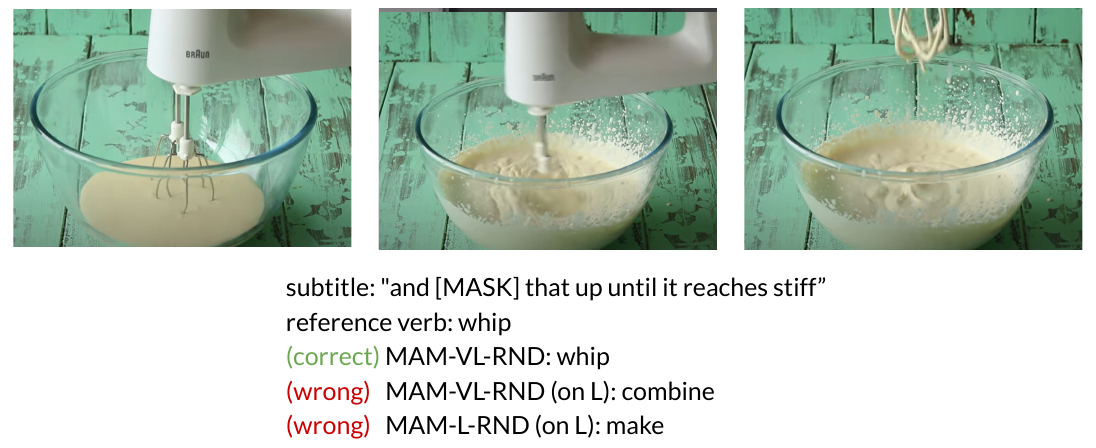}
    }
    \caption{Action inference task: \mamVTrandom is able to infer the correct action \word{whip} by reasoning upon the visual causal effect process. Under the language-only inference setting, models fail as the subtitle gives limited information.}
    \label{fig:verb_pred}
\end{figure*}

\begin{figure*}[h]
    \centering
    \resizebox{\textwidth}{!}{
    \includegraphics{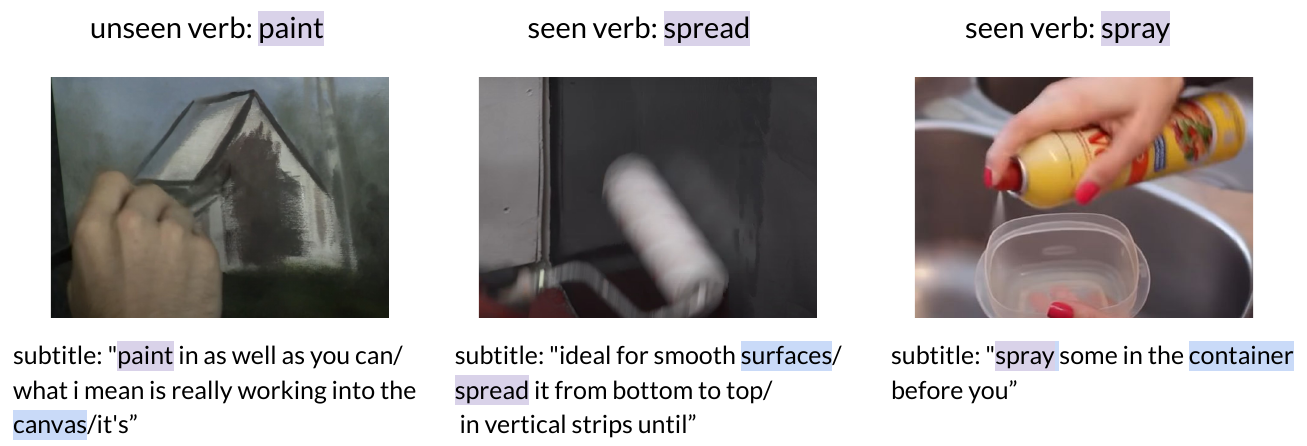}
    }
    \caption{Action generalization: \mamVTrandom generalizes to the unseen verb \word{paint}. Its causal effect is similar to the two seen verbs \word{spread} \& \word{spray} that produce a light coat on the surface of an object. For readability, we use \word{/} to indicate the sentence boundary. 
    }
    \label{fig:verb_generalization}
\end{figure*}

\begin{figure*}
    \centering
    \resizebox{\textwidth}{!}{\includegraphics{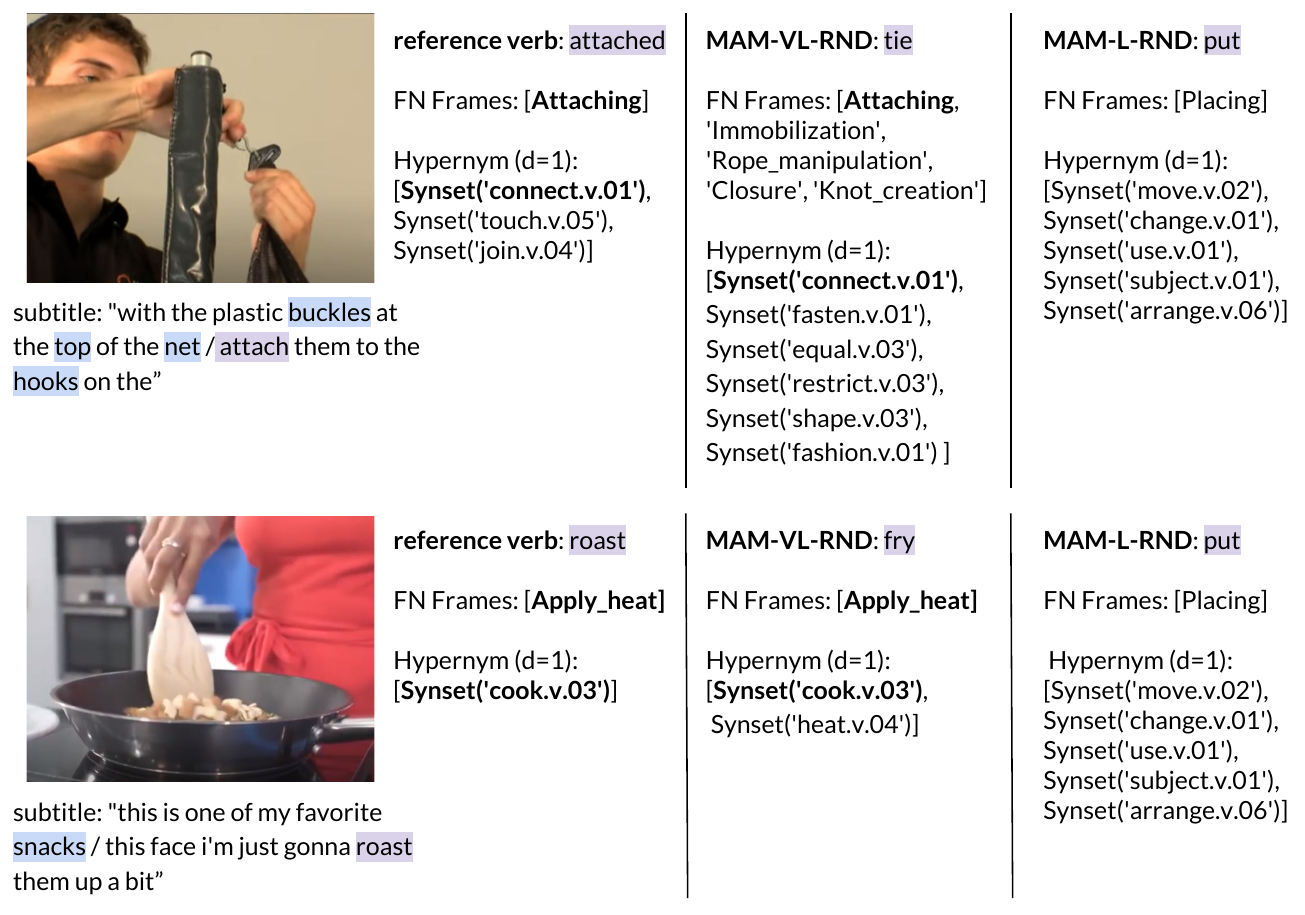}
    }
    \caption{Action generalization: both \mamVTrandom and \mamTrandom predict the wrong verb, but \mamVTrandom predicts the verbs that share the same FrameNet frame and Hypernym (depth=1) with the reference verbs. \mamTrandom tends to predict one of the most common seen verbs, \word{put}.
    }
    \label{fig:verb_generalization_analysis}
\end{figure*}

Figure~\ref{fig:verb_pred} shows an example of various models' prediction on \gls{map} task. Under limited textual context, \mamVTrandom is able to infer the action by reasoning upon the perceptual effect.
Figure~\ref{fig:verb_generalization} gives an example of \mamVTrandom's extrapolation ability of seen to unseen verbs in terms of visual effect similarities. Performing the \textit{paint} action on the object \textit{canvas}, \textit{spread} on \textit{surface}, and \textit{spray} on \textit{container} result in similar post-conditions, namely, the surface is covered by a layer.

\paragraph{Generalization Analysis.}
Figure~\ref{fig:verb_generalization_analysis} gives two wrong prediction cases of  \mamVTrandom and \mamTrandom, where we can see 
\mamVTrandom is able to predict a verb that captures relations with the reference verb from the perspective of situational and lexico-taxonomic semantics.

\subsection{\gls{mem}}
\label{app:mem}

\begin{table*}[t]
    \centering
    \begin{tabular}{@{~}l|r|r|r@{~}}
    \toprule 
    MEM Negative Sampling Strategies  & Randomized-set & Video-based-set & Object-based-set \\
      & (avg.neg=$618$) & (avg.neg=$8$) & (avg.neg=$452$) \\
    & Acc & Acc & Acc  \\
    \midrule
    Randomized-based & 87.74 & 70.4 & 87.91  \\ \midrule
    Video-based & 88.08 & \textbf{72.45} & 88.04  \\ \midrule
    Object-based & \textbf{88.57}  & 70.82 & \textbf{89.55} \\ \bottomrule 
    \end{tabular}
    \caption{Intrinsic evaluation of \gls{mem} during development on predicting the correct video \texttt{[AFT]} frame(s) among negatives constructed using various sampling schemes when tested on $10\%$ test set. \label{tab:dev_mep_results}}
\end{table*}


During the development stage performed on $10$\%~of the training data and test data, we explored three masking strategies on controlling the negative video frames. In addition, each candidate video frame has its corresponding video type: \texttt{[BEF]} for the pre-condition, \texttt{[ACT]} for the action process, and \texttt{[AFT]} for the post-condition:
\begin{itemize}
        \item \textbf{Randomized subclips}: randomized \texttt{[BEF]}, \texttt{[ACT]} and \texttt{[AFT]} subclips across video clips. 
        \item \textbf{Video-based subclips}: \texttt{[BEF]}, \texttt{[ACT]} and \texttt{[AFT]} subclips across multiple video clips that belong to the same \textbf{video id}.
        \item \textbf{Object-based subclips}: \texttt{[BEF]}, \texttt{[ACT]} and \texttt{[AFT]} subclips across multiple video clips that have the same object identified in the video clip. 
 \end{itemize}
 
 \paragraph{Details on Controlling the Negative Video Frame Samples.} In order to enhance the model's understanding in disentangling the pre-condition, the action process, and the post-condition in the visual space, we consider having the intra-video-clip video frames of types such as \texttt{[BEF]} (pre-condition) and \texttt{[ACT]} (post-condition) in the candidate set of the input. However, we exclude \texttt{[AFT]} video frames within the same clip as it becomes difficult and potentially unfair for the model to distinguish from due to its close temporal proximity.
 
\paragraph{Quantitative Results Across Different Negative Sampling Strategies During Development.}
As displayed in Table~\ref{tab:dev_mep_results}, the test set constructed based on a \textbf{video-based} sampling strategy is the most challenging one across three \gls{mem} models, despite that it yields the lowest average negative video frames in a batch (avg. neg is $8$). In general, when the negative samples in the test set are constructed identically as the train set, the corresponding trained model has the highest accuracy, e.g., the video-based model has $+2$pp acc over the randomized and object-based one. Moreover, both video-based and object-based \gls{mem} outperform than randomized one when evaluated on the randomized test set, emphasizing the importance of dedicated control on the negative sampling process.

\begin{figure*}
    \centering
    \resizebox{\textwidth}{!}{\includegraphics{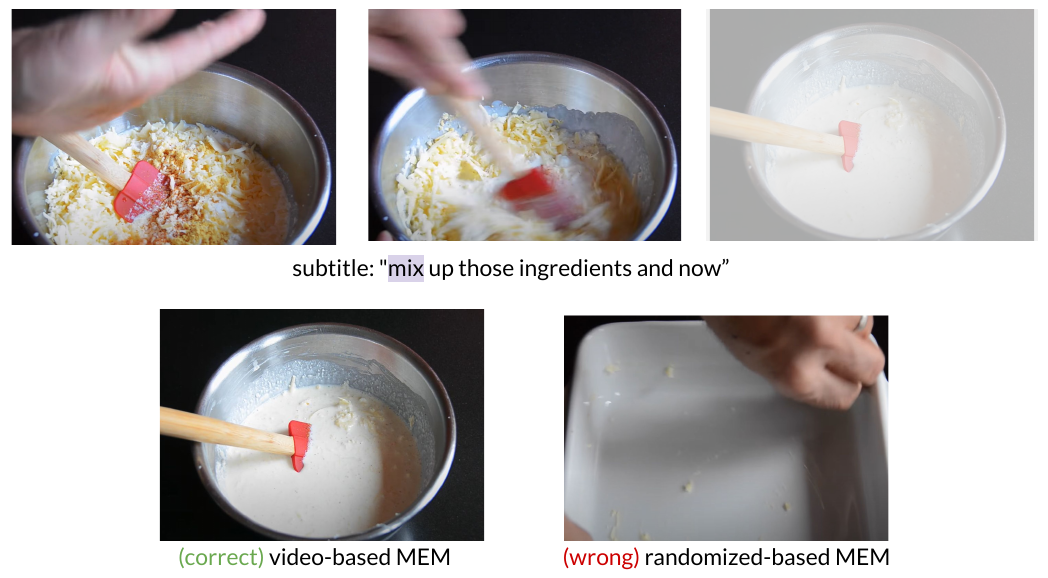}
    }
    \caption{Video-based \gls{mem} chose the correct  \texttt{[AFT]} frame, the randomized one fails. 
    }
    \label{fig:dev_mem_video_correct}
\end{figure*}

\paragraph{Qualitative Analysis During Development.}
As seen in Figure~\ref{fig:dev_mem_video_correct}, the video-based model is able to pick the correct post-condition for \textit{mixing the ingredients} while the randomized-based model struggles to find a correct video frame. We also probe into several failure modes and found some patterns. To begin, the candidate set that contains temporal close clips leads to wrong predictions more often than the temporal distant ones. Concretely, under the wrong predictions set, the average minimum temporal difference within the competitor set is $120$ secs, compared to $211$ secs for the correct predictions set. Similarly, the candidate set that has the same object occurrences or same action occurrences challenges the model to choose the right \texttt{[AFT]} video frame(s). In the wrong predictions set, $25.66$/$26.72\%$ of the candidate sets has the same object/action occurrences, compared to $20.98\%$/$18.58\%$  of that for the correct prediction group.

\begin{figure*}
    \centering
    \resizebox{\textwidth}{!}{\includegraphics{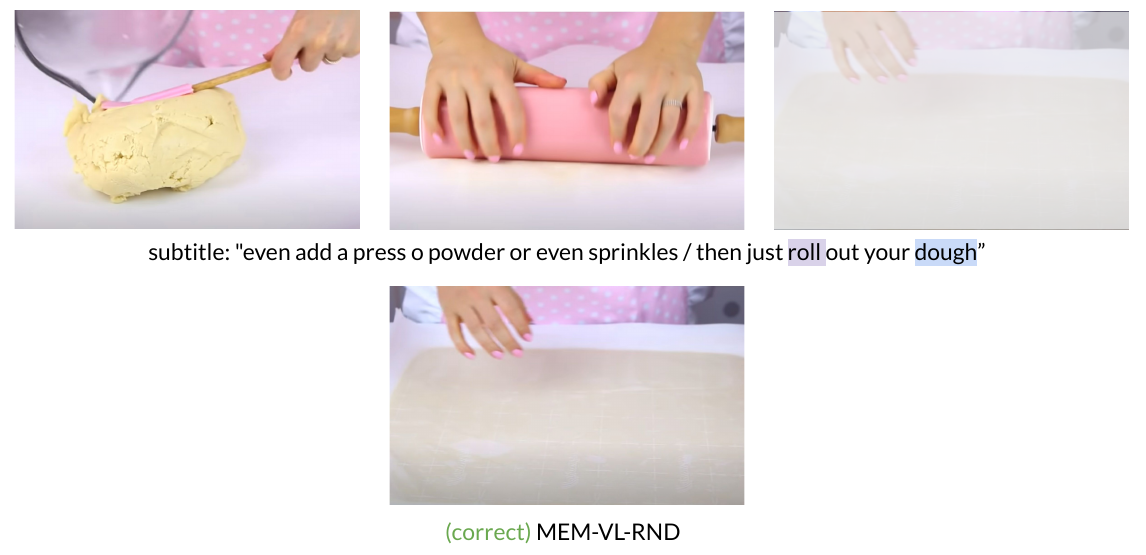}
    }
    \caption{Entity equivalence: although the action \word{roll} is unseen to \memVTrandom, it has successfully learned that \word{dough} could be rolled given its \word{foldability} which is a property shared among many seen objects like \word{cloth}.
    }
    \label{fig:mem_entity_equivalence}
\end{figure*}

\begin{figure*}
    \centering
    \resizebox{\textwidth}{!}{\includegraphics{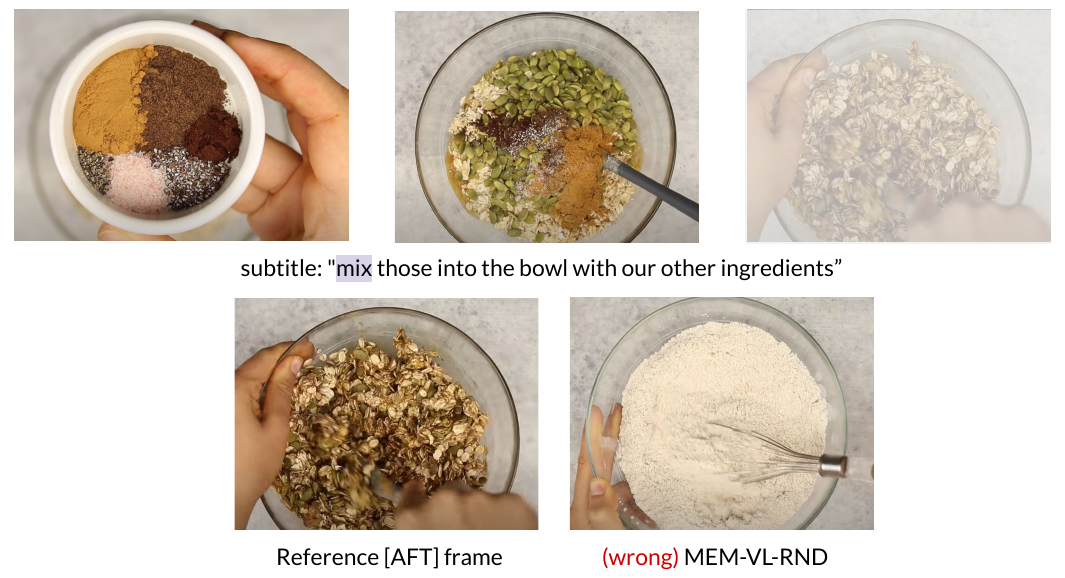}
    }
    \caption{Error analysis: \memVTrandom fails for a sensible reason, it selected the \texttt{[AFT]} frame that conceptually follows the instruction process of \word{mixing ingredients}. 
      }
    \label{fig:mem_video_incorrect}
\end{figure*}


\subsection{Pretraining Details}
\label{app:pretraining_details}
\paragraph{Hyperparameters.}
Following \citet{li-etal-2020-hero}, we used the AdamW optimizer \cite{adamW} with a learning rate of $3e-5$, weight decay of $0.01$, and warm-up steps of $10000$. 
We ran all the pretraining experiments  on 2 NVIDIA RTX A6000 GPUs with a batch size of $16$ per single GPU, and gradient accumulation steps \cite{DBLP:conf/wmt/OttEGA18} are set to $2$. We set the global training steps to $100000$ with maximum training time of $48$ hours.
\paragraph{Model Checkpoints.}
For model variants trained with either \gls{mam} or \gls{mem} pretraining tasks, the best model checkpoints are saved based on the highest task accuracy on the validation set.
Regarding the models that were with the joint task configuration, specifically \gls{multi}, the final model checkpoint is saved based on the best validation accuracy achieved on both tasks. 
Note that, for \gls{multi}-based models, each task undergoes training on only $50\%$ of the entire train set.
\paragraph{Ablated Models.}
We achieve modality ablation via inter-modal attention masking. More specifically, the weights of \mamTrandom are updated with the loss function:
    \begin{gather*}\label{eqn:mam_l}
        \mathcal{L}_{MAM}(\theta) = -\log P_{\theta}(\vb{s}_{a^*} | S_{\setminus \vb{s}_{a^*}})
    \end{gather*}

As for \memVrandom, we update its weights with the loss function:
    \begin{gather*}\label{eqn:mem_v}
        \mathcal{L}_{MEM}(\theta) = -\log P_{\theta}(\vb{v}_{\texttt{[AFT]}}| V_{\setminus \vb{v}_{\texttt{[AFT]}}})
    \end{gather*}

\subsection{PROST Task}
\label{app:prost}
\paragraph{Zero-shot Inference.}
The task is formulated as a cloze style task, where the \texttt{[MASK]} token should be reconstructed with an object in a set of $4$~mentioned candidates in the context that can afford the action.
Following \citet{aroca-ouellette-etal-2021-prost}, we obtain the logits of the $4$~object candidates instead of the whole vocabulary from the pretrained \gls{mam} head. Subsequently, we compute the probabilities of these candidates using the Softmax function. The object with the highest probability is the model's decision. 
\paragraph{PROST Probed Object Occurrences in \gls{cae}.}
As shown in Table.~\ref{tab:prost_action_object}, all the objects probed in the PROST affordance groups are seen to our models. Note that the statistic does not describe the occurrences of action-object combinations as only "Slide" is a seen verb to our models.
In terms of action--object combinations, only  [\prostafford{slide}, \prostobject{oil}] and [\prostafford{slide}, \prostobject{grease}], 
 are seen to our model. We hypothesize that \multiVT and \mamVT may have leveraged the shared visual properties of these objects, i.e.,~slipperiness, to extend the affordance understanding on the other objects like \prostobject{soap} and \prostobject{frost}.
 
\begin{table*}[ht]
\centering
\resizebox{0.95\textwidth}{!}{
	\begin{tabular}{@{~}l@{~}|lll@{~}}
		\toprule
		Affordance Concepts & Original & Inversed \\ 
		\midrule
            Stack & book: 481, block: 697, box: 1,632, coin: 113, plate: 1353 &  ball: 2,040, bottle: 918, egg: 3,083, flower: 1,068, lamp: 108 \\
            Roll &  apple: 690, ball: 2,040, bottle: 918, egg: 3,083, can: 522 &  book: 481, block: 697, box: 1,632, mirror: 168, microwave: 517 \\
            Grasp & ball: 2,040, block: 697, book: 481, bottle: 918, flower: 1,068 & flour: 2,692, rice: 1,233, salt: 2,862, snow: 151, sugar: 3227 \\
            Break & bottle: 918, egg: 3,083, glass: 1,254, mirror: 168, plate: 1,353 & ball: 2,040, coin: 113, pen: 252, pillow: 222, shirt: 373   \\
            \underline{Slide} & ice: 880, frost: 56, grease: 215, oil: 3,711, soap: 454 & carpet: 312, concrete: 246, grass: 229, gravel: 71, rubber: 133  \\
            Bounce & asphalt: 10, brick: 189, concrete: 246, rubber: 133, steel: 176 & carpet: 312, foam: 290, grass: 229, leave: 1012, snow: 151 \\            
  	\bottomrule
	\end{tabular}
 }
\caption{\protect Affordance concepts probed in the \gls{prost} and their corresponding object set, along with the total number of training instances per object in the \gls{cae} dataset.} \label{tab:prost_action_object}
\end{table*}

\paragraph{Model Robustness: Original vs. Inverses.}
The individual results on original and inverse templates across the concepts are shown in Table~\ref{tab:prost_template_result_breakdown}.

\begin{table*}[ht]
\captionsetup{justification=raggedleft}
\centering
\resizebox{0.95\textwidth}{!}{
    \begin{tabular}{@{~}l@{~}|rrrrrrrrrrrr@{~}}
		\toprule
		Model & \multicolumn{2}{c}{Stack}  & \multicolumn{2}{c}{Roll} & \multicolumn{2}{c}{Grasp} & \multicolumn{2}{c}{Break} &  \multicolumn{2}{c}{\underline{Slide}} & \multicolumn{2}{c}{Bounce}   \\ 
		     & Ori. & Inv.  & Ori. & Inv. & Ori. & Inv. & Ori. & Inv. &  Ori. & Inv. & Ori. & Inv.  \\ \midrule
            \mamT  & 0.0 & 60.0 & 30.0 & 10.0 & 8.0 & 40.0 & 34.0 & 20.0 & 44.0 & 4.0 & 6.0 & 40.0  \\
            \mamVT  & 20.0 & 24.0 & 10.0 & 40.0 & 2.0 & 64.0 & 32.0 & 20.0 & 40.0 & 20.0 & 20.0 & 26.0 \\
            \multiVT & 0.0 & 68.0 & 40.0 & 8.0 & 2.0 & 58.0 & 32.0 & 20.0 & 80.0 & 0.0 & 0.0 & 76.0\\
            \flava & 21.2 & 16.7 & 49.5 & 4.3 & 0.2 & 68.8 & 45.5 & 3.6 & 43.6 & 4.0 & 8.2 & 39.9 \\
            \roberta & 27.33 & 27.42 & 32.17 & 17.67 & 20.25 & 25.42 & 56.92 & 5.0 & 17.83 & 35.33 & 26.75 & 24.33 \\
  	\bottomrule
    \end{tabular}
    }
\caption{\protect Accuracy of original vs. inverse templates across affordance groups.} \label{tab:prost_template_result_breakdown}
\end{table*}

\end{document}